\documentclass[journal]{IEEEtran}
\usepackage{colortbl}
\usepackage{longtable}
\usepackage{indentfirst}
\usepackage{comment}
\usepackage{lineno,hyperref}
\usepackage{longtable}
\modulolinenumbers[5]
\usepackage{graphicx}
\usepackage{amsfonts}
\usepackage{verbatim}
\usepackage{amsmath}
\usepackage{algorithm}
\usepackage{caption}
\usepackage{algorithmicx}
\usepackage{algpseudocode}
\floatname{algorithm}{Algorithm}

\usepackage{booktabs}
\usepackage{makecell}

\newtheorem{myDef}{Definition}

\usepackage{bm}
\usepackage{epstopdf}

\usepackage{color}
\ifCLASSINFOpdf
\else
\fi
\begin{document}
%
\title{A Survey of Optimization Methods from \\ a Machine Learning Perspective}
%
%
%
\author{Shiliang~Sun, Zehui Cao, Han Zhu, and Jing~Zhao 

\thanks{This work was supported by NSFC Project 61370175 and Shanghai Sailing Program 17YF1404600.}
\thanks{Shiliang~Sun, Zehui Cao, Han Zhu, and Jing~Zhao are with School of Computer
Science and Technology, East China Normal University, 3663 North Zhongshan
Road, Shanghai 200062, P. R. China. E-mail: slsun@cs.ecnu.edu.cn, shiliangsun@gmail.com (Shiliang Sun); jzhao@cs.ecnu.edu.cn, jzhao2011@gmail.com (Jing Zhao)}
}


%
%

\markboth{}%
{Sun \MakeLowercase{\textit{et al.}}: A Survey of Optimization Methods from a Machine Learning Perspective}
%



\maketitle

\begin{abstract}
Machine learning develops rapidly, which has made many theoretical breakthroughs and is widely applied in various fields. Optimization, as an important part of machine learning, has attracted much attention of researchers. With the exponential growth of data amount and the increase of model complexity, optimization methods in machine learning face more and more challenges. A lot of work on solving optimization problems or improving optimization methods in machine learning has been proposed successively. The systematic retrospect and summary of the optimization methods from the perspective of machine learning are of great significance, which can offer guidance for both developments of optimization and machine learning research. In this paper, we first describe the optimization problems in machine learning. Then, we introduce the principles and progresses of commonly used optimization methods. Next, we summarize the applications and developments of optimization methods in some popular machine learning fields. Finally, we explore and give some challenges and open problems for the optimization in machine learning.
\end{abstract}

\begin{IEEEkeywords}
Machine learning, optimization method, deep neural network, reinforcement learning, approximate Bayesian inference.
\end{IEEEkeywords}

%
\IEEEpeerreviewmaketitle

\section{Introduction}\label{sec:introduction}

\IEEEPARstart{R}{ecently}, machine learning has grown at a remarkable rate, attracting a great number of researchers and practitioners. It has become one of the most popular research directions and plays a significant role in many fields, such as machine translation, speech recognition, image recognition, recommendation system, etc. Optimization is one of the core components of machine learning.
The essence of most machine learning algorithms is to build an optimization model and learn the parameters in the objective function from the given data. In the era of immense data, the effectiveness and efficiency of the numerical optimization algorithms dramatically influence the popularization and application of the machine learning models. In order to promote the development of machine learning, a series of effective optimization methods were put forward, which have improved the performance and efficiency of machine learning methods.

From the perspective of the gradient information in optimization, popular optimization methods can be divided into three categories: first-order optimization methods, which are represented by the widely used stochastic gradient methods; high-order optimization methods, in which Newton's method is a typical example; and heuristic derivative-free optimization methods, in which the coordinate descent method is a representative.
\par
As the representative of first-order optimization methods, the stochastic gradient descent method {\cite{robbins1951stochastic,jain2018parallelizing}}, as well as its variants, has been widely used in recent years and is evolving at a high speed. However, many users pay little attention to the characteristics or application scope of these methods. They often adopt them as black box optimizers, which may limit the functionality of the optimization methods. In this paper, we comprehensively introduce the fundamental optimization methods. Particularly, we systematically explain their advantages and disadvantages, their application scope, and the characteristics of their parameters. We hope that the targeted introduction will help users to choose the first-order optimization methods more conveniently and make parameter adjustment more reasonable in the learning process.

\par
Compared with first-order optimization methods, high-order methods {\cite{shanno1970conditioning,hu2019structured,pajarinen2019compatible}} converge at a faster speed in which the curvature information makes the search direction more effective. High-order optimizations attract widespread attention but face more challenges. The difficulty in high-order methods lies in the operation and storage of the inverse matrix of the Hessian matrix. To solve this problem, many variants based on Newton's method have been developed, most of which try to approximate the Hessian matrix through some techniques \cite{dennis1977quasi,martens2010deep}. In subsequent studies, the stochastic quasi-Newton method and its variants are introduced to extend high-order methods to large-scale data \cite{roosta2016sub,xu2016sub,bollapragada2016exact}.

\par
Derivative-free optimization methods {\cite{rios2013derivative,berahas2019derivative}} are mainly used in the case that the derivative of the objective function may not exist or be difficult to calculate. There are two main ideas in derivative-free optimization methods. One is adopting a heuristic search based on empirical rules, and the other is fitting the objective function with samples. Derivative-free optimization methods can also work in conjunction with gradient-based methods.

\par
Most machine learning problems, once formulated, can be solved as optimization problems. Optimization in the fields of deep neural network, reinforcement learning, meta learning, variational inference and Markov chain Monte Carlo encounters different difficulties and challenges. The optimization methods developed in the specific machine learning fields are different, which can be inspiring to the development of general optimization methods.
\par
Deep neural networks (DNNs) have shown great success in pattern recognition and machine learning. There are two very popular NNs, i.e., convolutional neural networks (CNNs) \cite{lecun1998gradient} and recurrent neural networks (RNNs), which play important roles in various fields of machine learning. CNNs are feedforward neural networks with convolution calculation. CNNs have been successfully used in many fields such as image processing \cite{krizhevsky2012imagenet,sermanet2013overfeat}, video processing \cite{karpathy2014large} and natural language processing (NLP) \cite{kim2014convolutional,ji20123d}. RNNs are a kind of sequential model and very active in NLP \cite{lai2015recurrent,cho2014learning, RNNFTC, SRWDRNN}. Besides, RNNs are also popular in the fields of image processing \cite{gregor2015draw,oord2016pixel} and video processing \cite{ullah2017action}. {In the field of constrained optimization, RNNs can achieve excellent results {\cite{xia2015bi,zhang2015complex,xia2018robust,xia2019two}.} In these works, the parameters of weights in RNNs can be learned by analytical methods, and these methods can find the optimal solution according to the trajectory of the state solution.
}
Stochastic gradient-based algorithms are widely used in deep neural networks \cite{duchi2011adaptive,zeiler2012adadelta,tieleman2012lecture,kingma2014adam}. However, various problems are emerging when employing stochastic gradient-based algorithms. For example, the learning rate will be oscillating in the later training stage of some adaptive methods \cite{reddi2018convergence,radford2015unsupervised}, which may lead to the problem of non-converging. Thus, further optimization algorithms based on variance reduction were proposed to improve the convergence rate \cite{roux2012stochastic,johnson2013accelerating}. Moreover, combining the stochastic gradient descent and the characteristics of its variants is a possible direction to improve the optimization. Especially, switching an adaptive algorithm to the stochastic gradient descent method can improve the accuracy and convergence speed of the algorithm \cite{keskar2017improving}.

\par
Reinforcement learning (RL) is a branch of machine learning, for which an agent interacts with the environment by trial-and-error mechanism and learns an optimal policy by maximizing cumulative rewards \cite{sutton1998reinforcement}. Deep reinforcement learning combines the RL and deep learning techniques, and enables the RL agent to have a good perception of its environment. Recent research has shown that deep learning can be applied to learn a useful representation for reinforcement learning problems \cite{mattner2012learn,mnih2013playing,mnih2015human,bengio2009learning,mousavi2016deep}. Stochastic optimization algorithms are commonly used in RL and deep RL models.

Meta learning \cite{schmidhuber1987evolutionary, Schaul2010}  has recently become very popular in the field of machine learning. The goal of meta learning is to design a model that can efficiently adapt to the new environment with as few samples as possible. The application of meta learning in supervised learning can solve the few-shot learning problems \cite{finn2017model}. In general, meta learning methods can be summarized into the following three types \cite{ModelvsOptimizationMetaLearning}: metric-based methods \cite{bromley1994signature, koch2015siamese, vinyals2016matching, snell2017prototypical}, model-based methods \cite{santoro2016meta, weston2014memory} and optimization-based methods \cite{andrychowicz2016learning, ravi2016optimization, finn2017model}.
We will describe the details of optimization-based meta learning methods in the subsequent sections.

\par
Variational inference is a useful approximation method which aims to approximate the posterior distributions in Bayesian machine learning. It can be considered as an optimization problem. For example, mean-field variational inference uses coordinate ascent to solve this optimization problem \cite{Bishop2006Pattern}. As the amount of data increases continuously, it is not friendly to use the traditional optimization method to handle the variational inference. Thus, the stochastic variational inference was proposed, which introduced natural gradients and extended the variational inference to large-scale data \cite{hoffman2013stochastic}.
\par
{Optimization methods have a significative influence on various fields of machine learning. For example, {\cite{pajarinen2019compatible}} proposed the transformer network using Adam optimization \cite{kingma2014adam}, which is applied to machine translation tasks.{ \cite{ledig2017photo}} proposed super-resolution generative adversarial network for image super resolution, which is also optimized by Adam. \cite{wu2017scalable} proposed Actor-Critic using trust region optimization to solve the deep reinforcement learning on Atari games as well as the MuJoCo environments.}

\par
The stochastic optimization method can also be applied to Markov chain Monte Carlo (MCMC) sampling to improve efficiency. In this kind of application, stochastic gradient Hamiltonian Monte Carlo (HMC) is a representative method \cite{chen2014stochastic} where the stochastic gradient accelerates the step of gradient update when handling large-scale samples. The noise introduced by the stochastic gradient can be characterized by introducing Gaussian noise and friction terms. Additionally, the deviation caused by HMC discretization can be eliminated by the friction term, and thus the Metropolis-Hasting step can be omitted. The hyper-parameter settings in the HMC will affect the performance of the model. There are some efficient ways to automatically adjust the hyperparameters and improve the performance of the sampler.

\par
The development of optimization brings a lot of contributions to the progress of machine learning. However, there are still many challenges and open problems for optimization problems in machine learning. 1) How to improve optimization performance with insufficient data in deep neural networks is a tricky problem. If there are not enough samples in the training of deep neural networks, it is prone to cause the problem of high variances and overfitting \cite{srivastava2014dropout}. In addition, non-convex optimization has been one of the difficulties in deep neural networks, which makes the optimization tend to get a locally optimal solution rather than the global optimal solution. 2) For sequential models, the samples are often truncated by batches when the sequence is too long, which will cause deviation. How to analyze the deviation of stochastic optimization in this case and correct it is vital. 3) The stochastic variational inference is graceful and practical, and it is probably a good choice to develop methods of applying high-order gradient information to stochastic variational inference. 4) It may be a great idea to introduce the stochastic technique to the conjugate gradient method to obtain an elegant and powerful optimization algorithm. The detailed techniques to make improvements in the stochastic conjugate gradient is an interesting and challenging problem.
\par

The purpose of this paper is to summarize and analyze classical and modern optimization methods from a machine learning perspective. The remainder of this paper is organized as follows.
Section \ref{sec:optimization} summarizes the machine learning problems from the perspective of optimization.
Section \ref{sec:fundamental} discusses the classical optimization algorithms and their latest developments in machine learning. Particularly, the recent popular optimization methods including the first and second order optimization algorithms are emphatically introduced.
Section \ref{sec:development} describes the developments and applications of optimization methods in some specific machine learning fields. Section \ref{sec:challenges} presents the challenges and open problems in the optimization methods. Finally, we conclude the whole paper.

\section{Machine Learning Formulated as Optimization}\label{sec:optimization}

Almost all machine learning algorithms can be formulated as an optimization problem to find the extremum of an objective function. Building models and constructing reasonable objective functions are the first step in machine learning methods. With the determined objective function, appropriate numerical or analytical optimization methods are usually used to solve the optimization problem.
\par
According to the modeling purpose and the problem to be solved, machine learning algorithms can be divided into supervised learning, semi-supervised learning, unsupervised learning, and reinforcement learning. Particularly, supervised learning is further divided into the classification problem (e.g., sentence classification \cite{kim2014convolutional,yin2016multichannel}, image classification \cite{yang2009linear,bazi2010gaussian,cirecsan2012multi}, etc.) and regression problem; unsupervised learning is divided into clustering and dimension reduction \cite{hartigan1979algorithm,guha2000rock,ding2002adaptive}, among others.

\subsection{Optimization Problems in Supervised Learning}
For supervised learning, the goal is to find an optimal mapping function $f (x)$ to minimize the loss function of the training samples,

\begin{equation}
	\min_{\theta}\frac{1}{N}\sum_{i=1}^N L(y^i,f(x^i,\theta)),
\end{equation}
\noindent
where $N$ is the number of training samples, $\theta$ is the parameter of the mapping function, $x^i$ is the feature vector of the $i$th samples, $y^i$ is the corresponding label, and $L$ is the loss function.

\par
There are many kinds of loss functions in supervised learning, such as the square of Euclidean distance, cross-entropy, contrast loss, hinge loss, information gain and so on. For regression problems, the simplest way is using the square of Euclidean distance as the loss function, that is, minimizing square errors on training samples. But the generalization performance of this kind of empirical loss is not necessarily good. Another typical form is structured risk minimization, whose representative method is the support vector machine. On the objective function, regularization items are usually added to alleviate overfitting, e.g., in terms of $L_2$ norm,

\begin{equation}
	\min_{\theta}\frac{1}{N}\sum_{i=1}^N L(y^i,f({x^i},\theta))+\lambda \parallel {\theta}\parallel _2^2.
\end{equation}
where $\lambda$ is the compromise parameter, which can be determined through cross-validation.

\subsection{Optimization Problems in Semi-supervised Learning}
	Semi-supervised learning (SSL) is the method between supervised and unsupervised learning, which incorporates labeled data and unlabeled data during the training process. It can deal with different tasks including classification tasks \cite{guillaumin2010multimodal,chapelle2005semi}, regression tasks \cite{zhou2005semi}, clustering tasks \cite{demiriz1999semi,kulis2009semi} and dimensionality reduction tasks \cite{zhang2007semi,chen2017semi}. There are different kinds of semi-supervised learning methods including
	self-training, generative models, semi-supervised support vector machines (S3VM) \cite{bennett1999semi}, graph-based methods, multi-learning method and others. We take S3VM as an example to introduce the optimization in semi-supervised learning.
	\par
	S3VM is a learning model that can deal with binary classification problems and only part of the training set in this problem is labeled. Let $D^l$ be labeled data which can be represented as $D^l=\{\{x^1,y^1\},\{x^2,y^2\},...,\{x^l,y^l\}\}$, and $D^u$ be unlabeled data which can be represented as $D^u=\{x^{l+1},x^{l+2},...,x^{N}\}$ with $N=l+u$. In order to use the information of unlabeled data, additional constraint on the unlabeled data is added to the original objective of SVM with slack variables $\zeta^i$. Specifically, define $\epsilon^j$ as the misclassification error of the unlabeled instance if its true label is positive and $z^j$ as the misclassification error of the unlabeled instance if its true label is negative. The constraint means to make $\sum_{j=l+1}^N \min (\epsilon^i,\zeta^i)$ as small as possible. Thus, an S3VM problem can be described as
	\begin{align}
	&\min \parallel \omega \parallel+C\left[ \sum_{i=1}^{l} \  \zeta^i+\sum_{j=l+1}^N \min (\epsilon^i,z^j)\right], \notag \\
	&\text{subject to }\notag \\
	&y^i(\textbf{w}\cdot x^i+b)+\zeta^i\ge 1 ,\zeta\ge 0,i=1,...,l \notag,\\
	&\textbf{w}\cdot x^j+b+\epsilon^j\ge 1 ,\epsilon\ge 0,j=l+1,...,N \notag, \\
	&-(\textbf{w}\cdot x^j+b)+z^j\ge 1 ,z^j\ge 0,
	\end{align}

\noindent where $C$ is a penalty coefficient. The optimization problem in S3VM is a mixed-integer problem which is difficult to deal with \cite{cheung2018optimization}.
There are various methods summarized in \cite{chapelle2008optimization} to deal with this problem, such as the branch and bound techniques \cite{BBSSSVM} and convex relaxation methods \cite{CSWLSVM}.

\subsection{Optimization Problems in Unsupervised Learning}
Clustering algorithms \cite{hartigan1979algorithm,murtagh1983survey,castro2000fast,ball1967clustering} divide a group of samples into multiple clusters ensuring that the differences between the samples in the same cluster are as small as possible, and samples in different clusters are as different as possible. The optimization problem for the $k$-means clustering algorithm is formulated as minimizing the following loss function:

\begin{equation}
	\min_S \sum_{k=1}^K\sum_{{x}\in S_k}\| {x}-\mu_k\|_2^2,
\end{equation}

\noindent
where $K$ is the number of clusters, ${x}$ is the feature vector of samples, $\mu_{k}$ is the center of cluster $k$, and $S_{k}$ is the sample set of cluster $k$. The implication of this objective function is to make the sum of variances of all clusters as small as possible.
\par
The dimensionality reduction algorithm ensures that the original information from data is retained as much as possible after projecting them into the low-dimensional space.
Principal component analysis (PCA) \cite{wold1987principal,jolliffe2011principal,tipping1999probabilistic} is a typical algorithm of dimensionality reduction methods.
The objective of PCA is formulated to minimize the reconstruction error as
\begin{equation}
	\min \sum_{i=1}^N\|{\overline x}^{i}-{x}^{i} \|_2^2 \quad  \mathrm{where} \quad {\overline{x}}^{i}=\sum_{j=1}^{D'}z_j^{i}{e}_j, D\gg D',
\end{equation}
where $N$ represents the number of samples, ${x}_i$ is a $D$-dimensional vector, ${\overline{x}}^{i}$ is the reconstruction of ${x}^{i}$. ${z}^{i}=\{z_1^{i},...,z_{D'}^{i}\}$ is the projection of ${x}^{i}$ in $D'$-dimensional coordinates. {${e}_j$ is the standard orthogonal basis under $D'$-dimensional coordinates.}

Another common optimization goal in probabilistic models is to find an optimal probability density function of $p({x})$, which maximizes the logarithmic likelihood function (MLE) of the training samples,
\begin{equation}
\max \sum_{i=1}^N\ln p({x^i};\theta).
\end{equation}
In the framework of Bayesian methods, some prior distributions are often assumed on parameter $\theta$, which also has the effect of alleviating overfitting.

\par
\subsection{Optimization Problems in Reinforcement Learning}
Reinforcement learning \cite{mnih2015human,sutton2018reinforcement,kaelbling1996reinforcement}, unlike supervised learning and unsupervised learning, aims to find an optimal strategy function, whose output varies with the environment. For a deterministic strategy, the mapping function from state $s$ to action $a$ is the learning target. For an uncertain strategy, the probability of executing each action is the learning target. In each state, the action is determined by $a=\pi(s)$, where $\pi(s)$ is the policy function.
\par
The optimization problem in reinforcement learning can be formulated as maximizing the cumulative return after executing a series of actions which are determined by the policy function,
\begin{equation}
	\max_\pi V_\pi (s) \quad \mathrm{where} \quad V_\pi (s)=\mathbb{E}\left[\sum_{k=0}^\infty \gamma^k r_{t+k}|S_t=s\right],
\end{equation}
\noindent
where $V_{\pi}(s)$ is the value function of state $s$ under policy $\pi$, $r$ is the reward, and $\gamma \in [0,1]$ is the discount factor.
\par
{ \subsection{Optimization for Machine Learning}}
Overall, the main steps of machine learning are to build a model hypothesis, define the objective function, and solve the maximum or minimum of the objective function to determine the parameters of the model. In these three vital steps, the first two steps are the modeling problems of machine learning, and the third step is to solve the desired model by optimization methods.

\section{Fundamental Optimization Methods and Progresses}\label{sec:fundamental}

From the perspective of gradient information, fundamental optimization methods can be divided into first-order optimization methods, high-order optimization methods and derivative-free optimization methods. These methods have a long history and are constantly evolving. They are progressing in many practical applications and have achieved good performance. Besides these fundamental methods, preconditioning is a useful technique for optimization methods. Applying reasonable preconditioning can reduce the number of iterations and obtain better spectral characteristics.
These technologies have been widely used in practice. For the convenience of researchers, we summarize the existing common optimization toolkits in a table at the end of this section.

\subsection{First-Order Methods}

In the field of machine learning, the most commonly used first-order optimization methods are mainly based on gradient descent. In this section, we introduce some of the representative algorithms along with the development of the gradient descent methods. At the same time, the classical alternating direction method of multipliers and the Frank-Wolfe method in numerical optimization are also introduced.

\subsubsection{Gradient Descent}
\par
The gradient descent method is the earliest and most common optimization method. The idea of the gradient descent method is that variables update iteratively in the (opposite) direction of the gradients of the objective function. The update is performed to gradually converge to the optimal value of the objective function. The learning rate $\eta$ determines the step size in each iteration, and thus influences the number of iterations to reach the optimal value \cite{ruder2016overview}.
\par
The steepest descent algorithm is a widely known algorithm. The idea is to select an appropriate search direction in each iteration so that the value of the objective function minimizes the fastest. Gradient descent and steepest descent are not the same, because the direction of the negative gradient does not always descend fastest. Gradient descent is an example of using the Euclidean norm in steepest descent \cite{boyd2004convex}.
\par
Next, we give the formal expression of gradient descent method. For a linear regression model, we assume that $f_{\theta}(x)$ is the function to be learned, $L(\theta)$ is the loss function, and $\theta$ is the parameter to be optimized. The goal is to minimize the loss function with
\begin{equation}
L(\theta)=\frac{1}{2N}\sum_{i=1}^{N}(y^{i}-f_{\theta}(x^{i}))^{2} \label{eq1},
\end{equation}
\begin{equation}
f_{\theta}(x)=\sum_{j=1}^{D}\theta_{j}x_{j},
\end{equation}
where $N$ is the number of training samples, $D$ is the number of input features, $x^i$ is an independent variable with $x^i=(x_1^i,...,x_D^i)$ for $i=1,...,N$ and $y^i$ is the target output. The gradient descent alternates the following two steps until it converges:

\begin{enumerate}[]
	\item Derive $L(\theta)$ for $\theta_j$ to get the gradient corresponding to each $\theta_j$:
	\begin{equation}
	\frac{\partial L(\theta)}{\partial {\theta}_{_j}}=-\frac{1}{N}\sum_{i=1}^{N}(y^{i}-f_{\theta}(x^i))x_{j}^{i}.
	\end{equation}
	
	\item Update each $\theta_j$ in the negative gradient direction to minimize the risk function:
	\begin{equation}
	{\theta}_{j}^{'}={\theta}_{j}+\eta \cdot \frac{1}{N}\sum_{i=1}^{N}(y^{i}-f_{\theta}(x^i))x_{j}^{i}.
	\end{equation}
\end{enumerate}

\par
The gradient descent method is simple to implement. The solution is global optimal when the objective function is convex.
It often converges at a slower speed if the variable is closer to the optimal solution, and more careful iterations need to be performed.
\par
In the above linear regression example, note that all the training data are used in each iteration step, so the gradient descent method is also called the batch gradient descent. If the number of samples is $N$ and the dimension of $x$ is $D$, the computation complexity for each iteration will be $O(ND)$. In order to mitigate the cost of computation, some parallelization methods were proposed \cite{Alspector1992A,nocedal2006numerical}. However, the cost is still hard to accept when dealing with large-scale data. Thus, the stochastic gradient descent method emerges.

\subsubsection{Stochastic Gradient Descent}
Since the batch gradient descent has high computational complexity in each iteration for large-scale data and does not allow online update, stochastic gradient descent (SGD) was proposed \cite{robbins1951stochastic}.
The idea of stochastic gradient descent is using one sample randomly to update the gradient per iteration, instead of directly calculating the exact value of the gradient. The stochastic gradient is an unbiased estimate of the real gradient \cite{robbins1951stochastic}.
The cost of the stochastic gradient descent algorithm is independent of sample numbers and can achieve sublinear convergence speed \cite{johnson2013accelerating}.
SGD reduces the update time for dealing with large numbers of samples and removes a certain amount of computational redundancy, which significantly accelerates the calculation.
In the strong convex problem, SGD can achieve the optimal convergence speed \cite{ nemirovsky1983problem, nemirovski2009robust, agarwal2009information, roux2012stochastic}.
Meanwhile, it overcomes the disadvantage of batch gradient descent that cannot be used for online learning.

\par
The loss function (\ref{eq1}) can be written as the following equation:
\begin{equation}
L(\theta)=\frac{1}{N}\sum_{i=1}^{N}\frac{1}{2}(y^{i}-f_{\theta}(x^i))^{2}=\frac{1}{N}\sum_{i=1}^{N}cost(\theta,(x^{i},y^{i})).
\end{equation}
If a random sample $i$ is selected in SGD, the loss function will be $L^*(\theta)$:
\begin{equation}
L^*(\theta)=cost(\theta,(x^{i},y^{i}))=\frac{1}{2}(y^{i}-f_{\theta}(x^i))^{2}.
\end{equation}
The gradient update in SGD uses the random sample $i$ rather than all samples in each iteration,
\begin{equation}
{\theta}^{'}={\theta}+\eta(y^{i}-f_{\theta}(x^i))x^{i}.
\end{equation}

\par
Since SGD uses only one sample per iteration, the computation complexity for each iteration is $O(D)$ where $D$ is the number of features.
The update rate for each iteration of SGD is much faster than that of batch gradient descent when the number of samples $N$ is large. SGD increases the overall optimization efficiency at the expense of more iterations, but the increased iteration number is insignificant compared with the high computation complexity caused by large numbers of samples.
It is possible to use only thousands of samples overall to get the optimal solution even when the sample size is hundreds of thousands. Therefore, compared with batch methods, SGD can effectively reduce the computational complexity and accelerate convergence.

\par
However, one problem in SGD is that the gradient direction oscillates because of additional noise introduced by random selection, and the search process is blind in the solution space.
Unlike batch gradient descent which always moves towards the optimal value along the negative direction of the gradient, the variance of gradients in SGD is large and the movement direction in SGD is biased. So, a compromise between the two methods, the mini-batch gradient descent method (MSGD), was proposed \cite{robbins1951stochastic}.

\par

The MSGD uses $b$ independent identically distributed samples ($b$ is generally in 50 to 256 \cite{ruder2016overview}) as the sample sets to update the parameters in each iteration. It reduces the variance of the gradients and makes the convergence more stable, which helps to improve the optimization speed. For brevity, we will call MSGD as SGD in the following sections.

\par
As a common feature of stochastic optimization, SGD has a better chance of finding the global optimal solution for complex problems.
The deterministic gradient in batch gradient descent may cause the objective function to fall into a local minimum for the multimodal problem. The fluctuation in the SGD helps the objective function jump to another possible minimum. However, the fluctuation in SGD always exists, which may more or less slow down the process of converging.

\par
There are still many details to be noted about the use of SGD in the concrete optimization process \cite{ruder2016overview}, such as the choice of a proper learning rate. A too small learning rate will result in a slower convergence rate, while a too large learning rate will hinder convergence, making loss function fluctuate at the minimum. One way to solve this problem is to set up a predefined list of learning rates or a certain threshold and adjust the learning rate during the learning process \cite{robbins1951textordfemininea, darken1992learning}. However, these lists or thresholds need to be defined in advance according to the characteristics of the dataset. It is also inappropriate to use the same learning rate for all parameters. If data are sparse and features occur at different frequencies, it is not expected to update the corresponding variables with the same learning rate. A higher learning rate is often expected for less frequently occurring features \cite{duchi2011adaptive, kingma2014adam}.

\par
Besides the learning rate, how to avoid the objective function being trapped in infinite numbers of the local minimum is a common challenge. Some work has proved that this difficulty does not come from the local minimum values, but comes from the ``saddle point'' \cite{sutskever2013training}. The slope of a saddle point is positive in one direction and negative in another direction, and gradient values in all directions are zero. It is an important problem for SGD to escape from these points. Some research about escaping from saddle points were developed \cite{allen2018natasha, ge2015escaping}.

\subsubsection{Nesterov Accelerated Gradient Descent}
\par

Although SGD is popular and widely used, its learning process is sometimes prolonged. How to adjust the learning rate, how to speed up the convergence, and how to prevent being trapped at a local minimum during the search are worthwhile research directions.

\par
Much work is presented to improve SGD. For example, the momentum idea was proposed to be applied in SGD \cite{polyak1964some}. The concept of momentum is derived from the mechanics of physics, which simulates the inertia of objects. The idea of applying momentum in SGD is to preserve the influence of the previous update direction on the next iteration to a certain degree. The momentum method can speed up the convergence when dealing with high curvature, small but consistent gradients, or noisy gradients \cite{goodfellow2016deep}. The momentum algorithm introduces the variable $v$ as the speed, which represents the direction and the rate of the parameter's movement in the parameter space. The speed is set as the average exponential decay of the negative gradient.

\par
In the gradient descent method, the speed update is $v=\eta\cdot(-\frac{\partial L(\theta)}{\partial (\theta)})$ each time. Using the momentum algorithm, the amount of the update $v$ is not just the amount of gradient descent calculated by $\eta\cdot(-\frac{\partial L(\theta)}{\partial (\theta)})$. It also takes into account the friction factor, which is represented as the previous update $v^{old}$ multiplied by a momentum factor ranging between [0, 1]. Generally, the mass of the object is set to 1. The formulation is expressed as
\begin{equation}
v=\eta\cdot(-\frac{\partial L(\theta)}{\partial (\theta)})+v^{old}\cdot mtm,
\end{equation}
where $mtm$ is the momentum factor. If the current gradient is parallel to the previous speed $v^{old}$, the previous speed can speed up this search. The proper momentum plays a role in accelerating the convergence when the learning rate is small. If the derivative decays to 0, it will continue to update $v$ to reach equilibrium and will be attenuated by friction. It is beneficial for escaping from the local minimum in the training process so that the search process can converge more quickly \cite{polyak1964some,sutskever2013importance}.
If the current gradient is opposite to the previous update $v^{old}$, the value $v^{old}$ will have a deceleration effect on this search.

\par
The momentum method with a proper momentum factor plays a positive role in reducing the oscillation of convergence when the learning rate is large. How to select the proper size of the momentum factor is also a problem. If the momentum factor is small, it is hard to obtain the effect of improving convergence speed. If the momentum factor is large, the current point may jump out of the optimal value point. Many experiments have empirically verified the most appropriate setting for the momentum factor is 0.9 \cite{ruder2016overview}.

\par
Nesterov Accelerated Gradient Descent (NAG) makes further improvement over the traditional momentum method \cite{sutskever2013importance, nesterov1983method}. In Nesterov momentum, the momentum $v^{old}\cdot mtm$ is added to $\theta$, denoted as $\widetilde{\theta}$. The gradient of $\widetilde{\theta}$ is used when updating. The detailed update formulae for parameters $\theta$ are as follows:

\par
\begin{equation}
\left\{
\begin{aligned}
\widetilde{\theta} &=\theta+v^{old}\cdot mtm, \\
v &=v^{old}\cdot mtm+\eta\cdot(-\frac{\partial L(\widetilde{\theta})}{\partial (\theta)}), \\
\theta' &=\theta+ v.
\end{aligned}
\right.
\end{equation}

The improvement of Nesterov momentum over momentum is reflected in updating the gradient of the future position instead of the current position. From the update formula, we can find that Nestorov momentum includes more gradient information compared with the traditional momentum method. Note that Nesterov momentum improves the convergence from $O(\frac{1}{k})$ (after $k$ steps) to $O(\frac{1}{k^{2}})$, when not using stochastic optimization \cite{nesterov1983method}.

\par
Another issue worth considering is how to determine the size of the learning rate. It is more likely to occur the oscillation if the search is closer to the optimal point. Thus, the learning rate should be adjusted. The learning rate decay factor $d$ is commonly used in the SGD's momentum method, which makes the learning rate decrease with the iteration period \cite{baird1999gradient}. The formula of the learning rate decay is defined as

\begin{equation}
\eta_{t} =\frac{\eta_{0}}{1 + d \cdot t},
\end{equation}

\noindent
where $\eta_{t}$ is the learning rate at the $t$th iteration, $\eta_{0}$ is the original learning rate, and $d$ is a decimal in $[0, 1]$.
As can be seen from the formula, the smaller the $d$ is, the slower the decay of the learning rate will be. The learning rate remains unchanged when $d = 0$ and the learning rate decays fastest when $d = 1$.

\subsubsection{Adaptive Learning Rate Method}
\par
The manually regulated learning rate greatly influences the effect of the SGD method. It is a tricky problem for setting an appropriate value of the learning rate \cite{duchi2011adaptive,kingma2014adam,darken1991note}.
Some adaptive methods were proposed to adjust the learning rate automatically. These methods are free of parameter adjustment, fast to converge, and often achieving not bad results. They are widely used in deep neural networks to deal with optimization problems.

\par
The most straightforward improvement to SGD is AdaGrad \cite{duchi2011adaptive}.
AdaGrad adjusts the learning rate dynamically based on the historical gradient in some previous iterations. The update formulae are as follows:

\begin{equation}
\left\{ \label{eq3.1.4}
\begin{aligned}
g_t &=\frac{\partial L(\theta_t)}{\partial \theta}, \\
V_t &={\sqrt{{\sum}_{i=1}^{t}(g_i)^2+\epsilon}}, \\
{\theta}_{t+1} &=\theta_t-\eta \frac{g_t}{V_t},
\end{aligned}
\right.
\end{equation}
where $g_t$ is the gradient of parameter $\theta$ at iteration $t$, $V_t$ is the accumulate historical gradient of parameter $\theta$ at iteration $t$, and ${\theta}_{t}$ is the value of parameter $\theta$ at iteration $t$.

The difference between AdaGrad and gradient descent is that during the parameter update process, the learning rate is no longer fixed, but is computed using all the historical gradients accumulated up to this iteration.
One main benefit of AdaGrad is that it eliminates the need to tune the learning rate manually. Most implementations use a default value of 0.01 for $\eta$ in (\ref{eq3.1.4}).

\par
Although AdaGrad adaptively adjusts the learning rate, it still has two issues. 1) The algorithm still needs to set the global learning rate $\eta$ manually.
2) As the training time increases, the accumulated gradient will become larger and larger, making the learning rate tend to zero, resulting in ineffective parameter update.

\par
AdaGrad was further improved to AdaDelta \cite{zeiler2012adadelta} and RMSProp \cite{tieleman2012lecture} for solving the problem that the learning rate will eventually go to zero. The idea is to consider not accumulating all historical gradients, but focusing only on the gradients in a window over a period, and using the exponential moving average to calculate the second-order cumulative momentum,

\begin{equation}
V_t=\sqrt{ \beta V_{t-1}+(1-\beta )(g_t)^2},
\end{equation}
where $\beta$ is the exponential decay parameter. Both RMSProp and AdaDelta have been developed independently around the same time, stemming from the need to resolve the radically diminishing learning rates of AdaGrad.

\par
Adaptive moment estimation (Adam) \cite{kingma2014adam} is another advanced SGD method, which introduces an adaptive learning rate for each parameter.
It combines the adaptive learning rate and momentum methods.
In addition to storing an exponentially decaying average of past squared gradients $V_t$, like AdaDelta and RMSProp, Adam also keeps an exponentially decaying average of past gradients $m_t$, similar to the momentum method:
{
\begin{equation}
m_t=\beta_1m_{t-1}+(1-\beta_1 )g_t,
\end{equation}
\begin{equation}
V_t=\sqrt {\beta_2V_{t-1}+(1-\beta_2 )(g_t)^2},
\end{equation}
}
\noindent
where $\beta_1$ and $\beta_2$ are exponential decay rates. The final update formula for the parameter $\theta$ is

\begin{equation}
{\theta}_{t+1}=m_t-\eta \frac{\sqrt{1-\beta_2}}{1-\beta_1}\frac{m_t}{{V_t}+\epsilon}.
\end{equation}
The default values of $\beta_1$, $\beta_2$, and $\epsilon$ are suggested to set to 0.9, 0.999, and $10^{-8}$, respectively. Adam works well in practice and compares favorably to other adaptive learning rate algorithms.

\subsubsection{Variance Reduction Methods}
\par
Due to a large amount of redundant information in the training samples, the SGD methods are very popular since they were proposed. However, the stochastic gradient method can only converge at a sublinear rate and the variance of gradient is often very large. How to reduce the variance and improve SGD to the linear convergence has always been an important problem.

\textbf{Stochastic Average Gradient}
The stochastic average gradient (SAG) method \cite{roux2012stochastic} is a variance reduction method proposed to improve the convergence speed.
The SAG algorithm maintains parameter $d$ recording the sum of the $N$ latest gradients $\{g_i\}$ in memory where $g_i$ is calculated using one sample $i, i \in \{1,...,N \}$. The detailed implementation is to select a sample $i_t$ to update $d$ randomly, and use $d$ to update the parameter $\theta$ in iteration $t$:
\begin{equation}
\left\{
\begin{aligned}
d &= d -\hat{g}_{i_t} + g_{i_t}(\theta_{t-1}), \\
\hat{g}_{i_t} &= g_{i_t}(\theta_{t-1}), \\
\theta_t &=\theta_{t-1}-\frac{\alpha}{N}d,
\end{aligned}
\right.
\end{equation}
where the updated item $d$ is calculated by replacing the old gradient $\hat{g}_{i_t}$ in $d$ with the new gradient $g_{i_t}(\theta_{t-1})$ in iteration $t$, { $\alpha$ is a constant representing the learning rate.}
Thus, each update only needs to calculate the gradient of one sample, not the gradients of all samples. The computational overhead is no different from SGD, but the memory overhead is much larger. This is a typical way of using space for saving time. The SAG has been shown to be a linear convergence algorithm \cite{roux2012stochastic}, which is much faster than SGD, and has great advantages over other stochastic gradient algorithms.
\par
However, the SAG method is only applicable to the case where the loss function is smooth and the objective function is convex \cite{roux2012stochastic, schmidt2017minimizing}, such as convex linear prediction problems.
In this case, the SAG achieves a faster convergence rate than the SGD. In addition, under some specific problems, it can even deliver better convergence than the standard batch gradient descent.
\par
\textbf{Stochastic Variance Reduction Gradient} Since the SAG method is only applicable to smooth and convex functions and needs to store the gradient of each sample, it is inconvenient to be applied in non-convex neural networks. The stochastic variance reduction gradient (SVRG) \cite{johnson2013accelerating} method was proposed to improve the performance of optimization in the complex models.
\par
The algorithm of SVRG maintains the interval average gradient $\tilde{\mu}$ by calculating the gradients of all samples in every $w$ iterations instead of in each iteration:
\begin{equation}
\tilde{\mu}=\frac{1}{N} \sum_{i=1}^{N}g_i(\tilde{\theta }),
\end{equation}
where $\tilde{\theta}$ is the interval update parameter. The interval parameter $\tilde{\mu}$ contains the average memory of all sample gradients in the past time for each time interval $w$.
SVRG picks uniform $i_t\in \{1,...,N \}$ randomly, and executes gradient updates to the current parameters:
\begin{equation}
\theta_t=\theta_{t-1}-\eta \cdot (g_{i_t}(\theta_{t-1})-g_{i_t}(\tilde{\theta})+\tilde{\mu}).
\end{equation}
The gradient is calculated up to two times in each update.
After $w$ iterations, perform $\tilde{\theta } \leftarrow \theta_w $ and start the next $w$ iterations. Through these update, $\theta_t$ and the interval update parameter $\tilde{\theta}$ will converge to the optimal $\theta^*$, and then $\tilde{\mu} \rightarrow 0$, and
\begin{equation}
g_{i_t}(\theta_{t-1})-g_{i_t}(\tilde{\theta})+\tilde{\mu} \rightarrow g_{i_t}(\theta_{t-1})-g_{i_t}({\theta}^*) \rightarrow 0.
\end{equation}

SVRG proposes a vital concept called variance reduction. This concept is related to the convergence analysis of SGD, in which it is necessary to assume that there is a constant upper bound for the variance of the gradients. This constant upper bound implies that the SGD cannot achieve linear convergence. However, in SVRG, the upper bound of variance can be continuously reduced due to the special update item $g_{i_t}(\theta_{t-1})-g_{i_t}(\tilde{\theta })+\tilde{\mu}$ , thus achieving linear convergence \cite{johnson2013accelerating}.

\par
The strategies of SAG and SVRG are related to variance reduction. Compared with SAG, SVRG does not need to maintain all gradients in memory, which means that memory resources are saved, and it can be applied to complex problems efficiently. Experiments have shown that the performance of SVRG is remarkable on a non-convex neural network \cite{johnson2013accelerating,allen2016variance,reddi2016stochastic}. There are also many variants of such linear convergence stochastic optimization algorithms, such as the SAGA algorithm \cite{defazio2014saga}.

\subsubsection{Alternating Direction Method of Multipliers}
\par
Augmented Lagrangian multiplier method is a common method to solve optimization problems with linear constraints. Compared with the naive Lagrangian multiplier method, it makes problems easier to solve by adding a penalty term to the objective. Consider the following example,
\begin{equation}
\min\left \{\theta_{1}(x)+\theta_{2}(y)|Ax+By=b,x\in \mathcal{X} ,y\in \mathcal{Y}\right \}.  \label{3.1.6}
\end{equation}
The augmented Lagrange function for problem (\ref{3.1.6}) is
\begin{equation}
\begin{aligned}
\mathcal{L}_{\beta}(x,y,\lambda )=&\theta_{1}(x)+\theta_{2}(y)-\lambda^{{\top}}(Ax+By-b)\\
&+\frac{\beta}{2}||Ax+By-b||^{2}.
\end{aligned}
\end{equation}
When solved by the augmented Lagrangian multiplier method, its $t$th step iteration starts from the given $\lambda_{t}$, and the optimization turns out to
{
\begin{equation}
\left\{ \label{eq3.1.6}
\begin{aligned}
(x_{t+1},y_{t+1})& = {\arg \min}\left \{\mathcal{L}_{\beta}(x,y,\lambda_{t} )|x\in \mathcal{X} ,y\in \mathcal{Y} \right \},\\
\lambda_{t+1}& = \lambda_{t}-\beta(Ax_{t+1}+By_{t+1}-b).
\end{aligned}
\right.
\end{equation} }

Separating the $(x,y)$ sub-problem in (\ref{eq3.1.6}), the augmented Lagrange multiplier method can be relaxed to the following alternating direction method of multipliers (ADMM) \cite{powell1969method,boyd2011distributed}. Its $t$th step iteration starts with the given $(y_{t},\lambda_{t})$, and the details of iterative optimization are as follows:
{
\begin{equation}
\left\{
\begin{aligned}
x_{t+1}& = {\arg \min}\left \{\theta_{1}(x)-(\lambda_{t})^{\top}Ax+\frac{\beta}{2}||\mathrm{Con}_x||^2|x\in \mathcal{X}  \right \},\\
y_{t+1}& = {\arg \min}\left \{\theta_{2}(y)-(\lambda_{t})^{\top}By+\frac{\beta}{2}||\mathrm{Con}_y||^{2}|y\in \mathcal{Y}  \right \},\\
\lambda_{t+1}& = \lambda_{t}-\beta(Ax_{t+1}+By_{t+1}-b),
\end{aligned}
\right.
\end{equation} }
where $\mathrm{Con}_x = Ax+By_{t}-b$ and $\mathrm{Con}_y = Ax_{t+1}+By-b$.
\par
The penalty parameter $\beta$ has a certain impact on the convergence rate of the ADMM. The larger $\beta$ is, the greater the penalties for the constraint term. In general, a monotonically increasing sequence of $\left \{ \beta_{t} \right \}$ can be adopted instead of the fixed $\beta$ \cite{nagurney1996transportation}. Specifically, an auto-adjustment criterion that automatically adjusts $\left \{ \beta_{t} \right \}$ based on the current value of $\left \{ x_{t} \right \}$ during the iteration was proposed, and applied for solving some convex optimization problems \cite{he2000alternating,hallac2017snapvx}.

\par
The ADMM method uses the separable operators in the convex optimization problem to divide a large problem into multiple small problems that can be solved in a distributed manner. In theory, the framework of ADMM can solve most of the large-scale optimization problems. However, there are still some problems in practical applications. For example, if we use a stop criterion to determine whether convergence occurs, the original residuals and dual residuals are both related to $\beta$, and $\beta$ with a large value will lead to difficulty in meeting the convergence conditions \cite{wahlberg2012admm}.

\subsubsection{Frank-Wolfe Method}
\par
In 1956, Frank and Wolfe proposed an algorithm for solving linear constraint problems \cite{Frank1956An}. The basic idea is to approximate the objective function with a linear function, then solve the linear programming to find the feasible descending direction, and finally make a one-dimensional search along the direction in the feasible domain. This method is also called the approximate linearization method.
\par
Here, we give a simple example of Frank-Wolfe method. Consider the optimization problem,
\begin{equation}
\left\{
\begin{array}{l}
\min \quad f(x),\\ \label{eq3.1.7.1}
\begin{aligned}
s.t. \quad Ax=b,\\
x\ge 0,
\end{aligned}
\end{array}
\right.
\end{equation}
where $A$ is an $m\times n$ full row rank matrix, and the feasible region is $S=\left \{ x|Ax=b,x \ge 0 \right \}$. Expand $f(x)$ linearly at $x_{0}$, $f(x)\approx f(x_{0})+\nabla f(x_{0})^{\top}(x-x_{0})$, and substitute it into equation (\ref{eq3.1.7.1}). Then we have
\begin{equation}
\left\{
\begin{array}{l}
\min \ \ f(x_{t})+\nabla f(x_{t})^{\top}(x-x_{t}),\\
\begin{aligned}
s.t. \ \ \ x\in S,
\end{aligned}
\end{array}
\right.
\end{equation}
which is equivalent to
\begin{equation}
\left\{ \label{eq3.1.7.2}
\begin{array}{l}
\min \ \ \nabla f(x_{t})^{\top}x,\\
\begin{aligned}
s.t. \ \ \ \  x\in S.
\end{aligned}
\end{array}
\right.
\end{equation}

Suppose there exist an optimal solution $y_{t}$, and then there must be
\begin{equation}
\left\{
\begin{aligned}
& \nabla f(x_{t})^{\top}y_{t} <\nabla f(x_{t})^{\top}x_{t},\\
& \nabla f(x_{t})^{\top}(y_{t}-x_{t}) <0.
\end{aligned}
\right.
\end{equation}
So $y_{t}-x_{t}$ is the decreasing direction of $f(x)$ at $x_{t}$. A fetch step of $\lambda_{t}$ updates the search point in a feasible direction. The detailed operation is shown in Algorithm \ref{3.1.7}.
\begin{algorithm}
	\caption{Frank-Wolfe Method \cite{Frank1956An, jaggi2013revisiting}}\label{3.1.7}
	\begin{algorithmic}
		\Require ~~$x_{0}$, $\varepsilon \ge 0$, $t:=0$
		\Ensure ~~$x^{*}$\\
		$y_t \leftarrow \min\nabla f(x_{t})^{\top}x$
		\While{$|\nabla f(x_{t})^{\top}(y_{t}-x_{t})|> \varepsilon $}
		\State $\lambda_{t} = \arg\min_{0\leq \lambda \leq 1} f(x_{t}+\lambda(y_{t}-x_{t}))$
		\State $x_{t+1}\approx x_{t}+\lambda_{t}(y_{t}-x_{t})$
		\State $t:=t+1$
		\State $y_t \leftarrow \min\nabla f(x_{t})^{\top}x$
		\EndWhile
		\State $x^{*}\approx x_{t}$
	\end{algorithmic}
\end{algorithm}
\par
The algorithm satisfies the following convergence theorem \cite{Frank1956An}:

\par
{ (1)} $x_{t}$ is the Kuhn-Tucker point of (\ref{eq3.1.7.1}) when $\nabla f(x_{t})^{\top}(y_{t}-x_{t}) = 0$.

\par
(2) Since $y_{t}$ is an optimal solution for problem (\ref{eq3.1.7.2}), the vector $d_t$ satisfies $d_{t}=y_{t}-x_{t}$ and is the feasible descending direction of $f$ at point $x_{t}$ when $\nabla f(x_{t})^{\top}(y_{t}-x_{t}) \neq 0$.
\par
The Frank-Wolfe algorithm is a first-order iterative method for solving convex optimization problems with constrained conditions. It consists of determining the feasible descent direction and calculating the search step size. The algorithm is characterized by fast convergence in early iterations and slower in later phases. When the iterative point is close to the optimal solution, the search direction and the gradient direction of the objective function tend to be orthogonal. Such a direction is not the best downward direction so that the Frank-Wolfe algorithm can be improved and extended in terms of the selection of the descending directions \cite{fukushima1984modified,patriksson2015traffic,clarkson2010coresets}.

\subsubsection{Summary}

We summarize the mentioned first-order optimization methods in terms of properties, advantages, and disadvantages in Table \ref{3.1.8}.

\begin{table*}[htbp]

	\caption{Summary of First-Order Optimization Methods }
	\begin{center}
		\begin{tabular}{p{2cm}p{4.5cm}p{4.5cm}p{4.5cm}}
			\toprule
			Method & Properties& Advantages & Disadvantages\\
			\midrule
			\label{3.1.8}
			& & \\
			GD &Solve the optimal value along the direction of the gradient descent. The method converges at a linear rate. & The solution is global optimal when the objective function is convex.&In each parameter update, gradients of total samples need to be calculated, so the calculation cost is high.\\ \hline
			& & \\
			SGD
			\cite{ robbins1951stochastic}
			 &The update parameters are calculated using a randomly sampled mini-batch. The method converges at a sublinear rate.& The calculation time for each update does not depend on the total number of training samples, and a lot of calculation cost is saved.&It is difficult to choose an appropriate learning rate, and using the same learning rate for all parameters is not appropriate. The solution may be trapped at the saddle point in some cases.\\ \hline
			& & \\
			NAG
			\cite{ nesterov1983method}
			&Accelerate the current gradient descent by accumulating the previous gradient as momentum and perform the gradient update process with momentum.& When the gradient direction changes, the momentum can slow the update speed and reduce the oscillation; when the gradient direction remains, the momentum can accelerate the parameter update.
			Momentum helps to jump out of locally optimal solution.
			&It is difficult to choose a suitable learning rate.\\ \hline
			& & \\
			AdaGrad
			\cite{ duchi2011adaptive}
			&The learning rate is adaptively adjusted according to the sum of the squares of all historical gradients.& In the early stage of training, the cumulative gradient is smaller, the learning rate is larger, and learning speed is faster.
			The method is suitable for dealing with sparse gradient problems.
			The learning rate of each parameter adjusts adaptively.
			&As the training time increases, the accumulated gradient will become larger and larger, making the learning rate tend to zero, resulting in ineffective parameter updates.
			A manual learning rate is still needed.
			It is not suitable for dealing with non-convex problems.
			\\ \hline
			& & \\
			AdaDelta/ RMSProp \cite{zeiler2012adadelta, tieleman2012lecture}
			&Change the way of total gradient accumulation to exponential moving average.& Improve the ineffective learning problem in the  late stage of AdaGrad.
			It is suitable for optimizing non-stationary and non-convex problems.
			&In the late training stage, the update process may be repeated around the local minimum.\\ \hline
			& & \\
			Adam
			\cite{ kingma2014adam}
			&
			Combine the adaptive methods and the momentum method. Use the first-order moment estimation and the second-order moment estimation of the gradient to dynamically adjust the learning rate of each parameter. Add the bias correction.& The gradient descent process is relatively stable.
			It is suitable for most non-convex optimization problems with large data sets and high dimensional space.
			&The method may not converge in some cases.\\ \hline
			& & \\
			SAG
			\cite { roux2012stochastic}
			 &The old gradient of each sample and the summation of gradients over all samples are maintained in memory. For each update, one sample is randomly selected and the gradient sum is recalculated and used as the update direction.& The method is a linear convergence algorithm, which is much faster than SGD.&The method is only applicable to smooth and convex functions and needs to store the gradient of each sample. It is inconvenient to be applied in non-convex neural networks.\\ \hline
			& & \\
			SVRG
			\cite{ johnson2013accelerating}
			&Instead of saving the gradient of each sample, the average gradient is saved at regular intervals. The gradient sum is updated at each iteration by calculating the gradients with respect to the old parameters and the current parameters for the randomly selected samples.& The method does not need to maintain all gradients in memory, which saves memory resources. It is a linear convergence algorithm.&To apply it to larger/deeper neural nets whose training cost is a critical issue, further investigation is still needed.\\
			\hline
			& & \\
			ADMM \cite{mairal2014spams}&
			The method solves optimization problems with linear constraints by adding a penalty term to the objective and separating variables into sub-problems which can be solved iteratively.& The method uses the separable operators in the convex optimization problem to divide a large problem into multiple small problems that can be solved in a distributed manner. The framework is practical in most large-scale optimization problems.&The original residuals and dual residuals are both related to the penalty parameter whose value is difficult to determine.\\ \hline
			& & \\
			Frank-Wolfe
			\cite{ Frank1956An}
			 &The method approximates the objective function with a linear function, solves the linear programming to find the feasible descending direction, and makes a one-dimensional search along the direction in the feasible domain.& The method can solve optimization problems with linear constraints, whose convergence speed is fast in early iterations. &The method converges slowly in later phases. When the iterative point is close to the optimal solution, the search direction and the gradient of the objective function tend to be orthogonal. Such a direction is not the best downward direction.\\ \hline
		\end{tabular}
	\end{center}
	\vspace{-0.5cm}

\end{table*}

\subsection{High-Order Methods}
The second-order methods can be used for addressing the problem where an objective function is highly non-linear and ill-conditioned.  They work effectively by introducing curvature information.
\par
This section begins with introducing the conjugate gradient method, which is a method that only needs first-order derivative information for well-defined quadratic
programming, but overcomes the shortcoming of the steepest descent method, and avoids the disadvantages of Newton's method of storing and calculating the inverse Hessian matrix. But note that when applying it to general optimization problems, the second-order gradient is needed to get an approximation to quadratic programming.
Then, the classical quasi-Newton method using second-order information is described. Although the convergence of the algorithm can be guaranteed, the computational process is costly and thus rarely used for solving large machine learning problems. In recent years, with the continuous improvement of high-order optimization methods, more and more high-order methods have been proposed to handle large-scale data by using stochastic techniques \cite{schraudolph2007stochastic,byrd2016stochastic,moritz2016linearly}. From this perspective, we discuss several high-order methods including the stochastic quasi-Newton method (integrating the second-order information and the stochastic method) and their variants. These algorithms allow us to use high-order methods to process large-scale data.
\par
\subsubsection{Conjugate Gradient Method}
The conjugate gradient (CG) approach is a very interesting optimization method, which is one of the most effective methods for solving large-scale linear systems of equations. It can also be used for solving nonlinear optimization problems \cite{nocedal2006numerical}.
As we know, the first-order methods are simple but have a slow convergence speed, and the second-order methods need a lot of resources. Conjugate gradient optimization is an intermediate algorithm, which can only utilize the first-order information for some problems but ensures the convergence speed like high-order methods.
\par
Early in the 1960s, a conjugate gradient method for solving a linear system was proposed, which is an alternative to Gaussian elimination \cite{hestenes1952methods}. Then in 1964, the conjugate gradient method was extended to handle nonlinear optimization for general functions \cite{nocedal2006numerical}. For years, many different algorithms have been presented based on this method, some of which have been widely used in practice. The main features of these algorithms are that they have faster convergence speed than steepest descent. Next, we describe the conjugate gradient method.
\par
Consider a linear system,
\begin{equation}
A\theta=b, \label{3.1.1}
\end{equation}
where $A$ is an $n\times n$ symmetric, positive-definite matrix. The matrix $A$ and vector $b$ are known, and we need to solve the value of $\theta$. The problem (\ref{3.1.1}) can also be considered as an optimization problem that minimizes the quadratic positive definite function,
\begin{equation}
\operatorname*{min}\limits_{\theta}\ \ F(\theta)=\frac{1}{2}\theta^{\top}  A\theta-b\theta+c.
\end{equation} 
The above two equations have an identical unique solution. It enables us to regard the conjugate gradient as a method for solving optimization problems.
\par
The gradient of $F(\theta)$ can be obtained by simple calculation, and it equals the residual of the linear system \cite{nocedal2006numerical}:
$
r(\theta)= \nabla F(\theta)=A\theta-b.
$

\begin{myDef}
	Conjugate: Given an $n\times n$ symmetric positive-definite matrix $A$, two non-zero vector $d_i,d_j$ are conjugate with respect to $A$ if
	
	\begin{equation}
	d_i^{\top}  Ad_j=0.
	\end{equation}
\end{myDef}
A set of non-zero vector $\{d_1, d_2,d_3,....,d_n\}$ is said to be conjugate { with respect to $A$} if any two unequal vectors are conjugate with respect to $A$ \cite{nocedal2006numerical}.

\par
Next, we introduce the detailed derivation of the conjugate gradient method. $\theta_{0}$ is a starting point, $\{d_t\}_{t=1}^{n-1}$ is a set of conjugate directions. In general, one can generate the update sequence $\{\theta_{1},\theta_{2},....,\theta_{n}\}$ by a iteration formula:
\begin{equation}
\theta_{t+1}=\theta_{t}+\eta_td_{t}.
\end{equation}
The step size $\eta_t$ can be
 obtained by a linear search, which means choosing $\eta_t$ to minimize the object function $f(\cdot)$ along $\theta_{t}+\eta_t d_t$.
 	After some calculations (more details in \cite{nocedal2006numerical,shewchuk1994introduction}), the update formula of $\eta_t$ is
 	\begin{equation}
 	\eta_t=\frac{r_t^{\top} r_t}{d_t^{\top} Ad_t}.
 	\end{equation}
The search direction $d_t$ is obtained by a linear combination of the negative
residual and the previous search direction,
\begin{equation}
d_t=-r_{t}+\beta_{t}d_{t-1},\label{3.1.2}
\end{equation}
where $r_{t}$ can be updated by $r_{t}=r_{t-1}+\eta_{t-1} A d_{t-1}$. The scalar $\beta_{t}$ is the update parameter, which can be determined by satisfying the requirement that $d_t$ and $d_{t-1}$
are conjugate with respect to $A$, i.e., $ d_t^{\top}Ad_{t-1}=0$. Multiplying both sides of the equation (\ref{3.1.2}) by $d_{t-1}^{\top} A$, one can obtain $\beta_t$ by
\begin{equation}
\beta_{t}=\frac{d_{t-1}^{\top} Ar_{t}}{d_{t-1}^{\top} Ad_{t-1}}.
\end{equation}
After several derivations of the above formula according to \cite{nocedal2006numerical}, the simplified version of $\beta_t$ is
\begin{equation}
\beta_{t}=\frac{r_{t}^{\top} r_{t}}{r_{t-1}^{\top} r_{t-1}}.
\end{equation}

The CG method, has a graceful property that generating a new vector $d_t$ only using the previous vector $d_{t-1}$, which does not need to know all the previous vectors $d_0,d_1,d_2\dots d_{t-2}$. The linear conjugate gradient algorithm is shown in Algorithm \ref{CGM}.

\begin{algorithm}
	\caption{Conjugate Gradient Method \cite{shewchuk1994introduction}} \label{CGM}
	\begin{algorithmic}
		\Require ~~$A$, $b$, $\theta_0$
		\Ensure ~~The solution $\theta^*$
        \State $r_0=A\theta_0-b$
		\State $d_{0}=-r_0$, $t=0$
		\While{Unsatisfied convergence condition}
		\State $ \eta_t=\frac{r_t^{\top} r_t}{d_t^{\top} Ad_t}$
		\State $\theta_{t+1}=\theta_t+\eta_t d_t$
		\State $r_{t+1}=r_t+\eta_t A d_t$
		\State $\beta_{t+1}=\frac{r_{t+1}^{\top} r_{t+1}}{r_{t}^{\top} r_{t}}$
		\State $d_{t+1}=-r_{t+1}+\beta_{t+1}d_t$
		\State $t=t+1$
		\EndWhile

	\end{algorithmic}
\end{algorithm}

\subsubsection{Quasi-Newton Methods}\label{sec:quasi-newton-method}
Gradient descent employs first-order information, but its convergence rate is slow. Thus, the natural idea is to use second-order information, e.g., Newton's method \cite{avriel2003nonlinear}. The basic idea of Newton's method is to use both the first-order derivative (gradient) and second-order derivative (Hessian matrix) to approximate the objective function with a quadratic function, and then solve the minimum optimization of the quadratic function. This process is repeated until the updated variable converges.
\par
The one-dimensional Newton's iteration formula is shown as
\par
\begin{equation}
\theta_{t+1}=\theta_{t}-\frac{f'(\theta_t)}{f''(\theta_t)},
\end{equation}
where $f$ is the object function. More general, the high-dimensional Newton's iteration formula is
\begin{equation}
\theta_{t+1}=\theta_t-\nabla^2f(\theta_t)^{-1}\nabla f(\theta_t)\ , \quad t\ge 0,
\end{equation}
where $\nabla^2f$ is a Hessian matrix of $f$. More precisely, if the learning rate (step size factor) is introduced, the iteration formula is shown as
\par

\begin{align}
&d_{t}=-\nabla^2f(\theta_t)^{-1}\nabla f(\theta_t),\notag \\
&\theta_{t+1}=\theta_{t}+\eta_td_{t},
\end{align}
where $d_{t}$ is the Newton's direction, $\eta_t$ is the step size. This method can be called damping Newton's method \cite{harker1990damped}. Geometrically speaking, Newton's method is to fit the local surface of the current position with a quadratic surface, while the gradient descent method is to fit the current local surface with a plane \cite{ayala1997combined}.

\textbf{Quasi-Newton Method}
Newton's method is an iterative algorithm that requires the computation of the inverse Hessian matrix of the objective function at each step, which makes the storage and computation very expensive. To overcome the expensive storage and computation, an approximate algorithm was considered which is called the quasi-Newton method. The essential idea of the quasi-Newton method is to use a positive definite matrix to approximate the inverse of the Hessian matrix, thus simplifying the complexity of the operation.
The quasi-Newton method is one of the most effective methods for solving non-linear optimization problems. Moreover, the second-order gradient is not directly needed in the quasi-Newton method, so it is sometimes more efficient than Newton's method. In the following section, we will introduce several quasi-Newton methods, in which the Hessian matrix and its inverse matrix are approximated by different methods.

\par
\textbf{Quasi-Newton Condition}
We first introduce the quasi-Newton condition. Assuming that the objective function $f$ can be approximated by a quadratic function, we can extend $f(\theta)$ to Taylor series at $\theta=\theta_{t+1}$, i.e.,
\begin{align}
f(\theta)&\approx f(\theta_{t+1})+\nabla f(\theta_{t+1})^{\top} (\theta-\theta_{t+1})\notag \\
&\quad+\frac{1}{2}(\theta-\theta_{t+1})^{\top} \nabla^2f(\theta_{t+1})(\theta-\theta_{t+1}).
\end{align}
Then we can compute the gradient on both sides of the above equation, and obtain
\begin{equation}
\nabla f(\theta)\approx \nabla f(\theta_{t+1})+\nabla^2 f(\theta_{t+1})(\theta-\theta_{t+1}) \label{3.2.2}.
\end{equation}
Set $\theta=\theta_t$ in (\ref{3.2.2}), we have
\begin{equation}
\nabla f(\theta_t)\approx \nabla f(\theta_{t+1})+\nabla^2 f(\theta_{t+1})(\theta_t-\theta_{t+1}).
\end{equation}
Use $B$ to represent the approximate matrix of the Hessian matrix. Set $s_t=\theta_{t+1}-\theta_t$, and $u_t=\nabla f(\theta_{t+1})-\nabla f(\theta_t)$. The matrix $B_{t+1}$ is satisfied that

\begin{equation}
u_t=B_{t+1}s_t.
\end{equation}
 This equation is called the quasi-Newton condition, or secant equation.

The search direction of quasi-Newton method is
\begin{equation}
d_t=-B_t^{-1}g_t,
\end{equation}
where $g_t$ is the gradient of $f$, and the update of quasi-Newton is
\begin{equation}
\theta_{t+1}=\theta_{t}+\eta_t d_t.
\end{equation}
The step size $\eta_t$ is chosen to satisfy the Wolfe conditions, which is a set of inequalities for inexact line searches $\min_{\eta_t} f(\theta_t+\eta_td_t)$ \cite{raydan1997barzilai}. Unlike Newton's method, quasi-Newton method uses $B_{t}$ to approximate the true Hessian matrix.
In the following paragraphs, we will introduce some particular quasi-Newton methods, in which $H_{t}$ is used to express the inverse of $B_t$, i.e., $H_t=B_t^{-1}$.
\par
\textbf{DFP}
In the 1950s, a physical scientist, William C. Davidon \cite{davidon1991variable}, proposed a new approach to solve nonlinear problems. Then Fletcher and Powel \cite{fletcher1963rapidly} explained and improved this method, which sparked a lot of research in the late 1960s and early 1970s \cite{dennis1977quasi}. DFP is the first quasi-Newton method named after the initials of their three names. The DFP correction formula is one of the most creative inventions in the field of non-linear optimization, shown as below:
\begin{equation}
B_{t+1}^{(DFP)}=(I-\frac{u_ts_t^{\top} }{u_t^{\top} s_t})B_t(I-\frac{s_tu_t^{\top} }{u_t^{\top} s_t})+\frac{u_tu_t^{\top} }{u_t^{\top} s_t}.
\end{equation}
The update formula of $H_{t+1}$ is
\begin{equation}
H_{t+1}^{DFP}=H_t-\frac{H_tu_tu_t^{\top} H_t}{u_t^{\top} H_tu_t}+\frac{s_ts_t^{\top} }{u_t^{\top} s_t}.
\end{equation}
\par
\textbf{BFGS}
Broyden, Fletcher, Goldfarb and Shanno proposed the BFGS method \cite{broyden1970convergence,fletcher1970new,goldfarb1970family,shanno1970conditioning}, in which $B_{t+1}$ is updated according to
\begin{equation}
B_{t+1}^{(BFGS)}=B_t-\frac{B_ts_ts_t^{\top} B_t}{s_t^{\top} B_ts_t}+\frac{u_tu_t^{\top} }{u_t^{\top} s_t}.
\end{equation}
The corresponding update of $H_{t+1}$ is
\begin{equation}
H_{t+1}^{(BFGS)}=(I-\frac{s_tu_t^{\top} }{s_t^{\top}u_t})H_{t}(I-\frac{u_ts_t^{\top} }{s_t^{\top} u_t})+\frac{u_ts_t^{\top} }{s_t^{\top} u_t}.
\end{equation}
\par
The quasi-Newton algorithm still cannot solve large-scale data optimization problem because the method generates a sequence of matrices to approximate the Hessian matrix. Storing these matrices needs to consume computer resources, especially for high-dimensional problems. It is also impossible to retain these matrices in the high-speed storage of computers, restricting its use to even small and midsize problems \cite{nocedal1980updating}.
\par
\textbf{L-BFGS}
Limited memory quasi-Newton methods, named L-BFGS  \cite{nocedal1980updating,liu1989limited} is an improvement based on the quasi-Newton method, which is feasible in dealing with the high-dimensional situation. The method stores just a few $n$-dimensional vectors, instead of retaining and computing fully dense $n\times n$ approximations of the Hessian \cite{sun2006optimization}.
The basic idea of L-BFGS is to store the vector sequence in the calculation of approximation $H_{t+1}$, instead of storing complete matrix $H_t$. L-BFGS makes further consolidation for the update formula of $H_{t+1}$,

\begin{align}
H_{t+1}&=(I-\frac{s_tu_t^{\top} }{u_t^{\top} s_t})H_{t}(I-\frac{u_ts_t^{\top} }{u_t^{\top} s_t})+\frac{s_ts_t^{\top} }{u_t^{\top} s_t}\notag \\
&=V_t^{\top} H_tV_t+\rho s_ts_t^{\top},
\end{align}
where
\begin{equation}
V_t=I-\rho u_ts_t^{\top},\qquad \rho_t=\frac{1}{s_t^{\top} u_t}.
\end{equation}
\par
The above equation means that the inverse Hessian approximation $H_{t+1}$ can be obtained using the sequence pair $\{s_l,u_l\}_{l=t-p+1}^{t}$. $H_{t+1}$ can be computed if we know pairs $\{s_l,y_l\}_{l=t-p+1}^{t}$. In other words, instead of storing and calculating the complete matrix $H_{t+1}$, L-BFGS only computes the latest $p$ pairs of $\{s_l,y_l\}$. According to the equation, a recursive procedure can be reached. When the latest $p$ steps are retained, the calculation of $H_{t+1}$ can be expressed as \cite{liu1989limited}
\begin{eqnarray}
&&H_{t+1}=(V_t^{\top} V_{t-1}^{\top} \cdots V_{t-p+1}^{\top} )H_t^0(V_{t-p+1}V_{t-p+2}\cdots V_t)\nonumber\\\nonumber
	&+&\rho_{t-p+1}(V_t^{\top} V_{t-1}^{\top} \cdots V_{t-p+2})s_{t-p+1}s_{t-p+1}^{\top} (V_{t-p+2}\cdots V_t)\\\nonumber
	&+&\rho_{t-p+2}(V_t^{\top} V_{t-1}^{\top} \cdot V_{t-p+3}^{\top} )s_{t-p+2}s_{t-p+2}^{\top} (V_{t-p+3}\cdots V_t)\\\nonumber
	&+&\cdots\\\nonumber
	&+&\rho_ts_ts_t^{\top}. \\
\end{eqnarray}
The update direction $d_t=H_tg_t$ can be calculated, where $g_t$ is the gradient of the objective function $f$. The detailed algorithm is shown in Algorithms \ref{TL} and \ref{L-BFGS}.
\par
\begin{algorithm}
	\caption{Two-Loop Recursion for $H_tg_t$ \cite{nocedal2006numerical}}\label{TL}
	\begin{algorithmic}
		\Require ~~$\nabla f_t$, $u_t$, $s_t$
		\Ensure ~~$H_{t+1}g_{t+1}$\\
		$g_t=\nabla f_t$\\
		$H_t^0=\frac{s_t^{\top} u_t}{\parallel u_t \parallel^2}\mathrm{I}$
		\For{$l=t-1$ to $t-p$}
		\State $\eta_l=\rho_ls_l^{\top}g_{l+1}$
		\State $g_l=g_{l+1}-\eta_l u_l$
		\EndFor
		\State $r_{t-p-1}=H_t^0g_{t-p}$
		\For{$l=t-p$ to $t-1$}
		\State$\beta_l=\rho_l u_l^{\top}\rho_{l-1}$
		\State$\rho_l=\rho_{l-1}+s_l(\eta_l-\beta_l)$
		\EndFor\\
        $H_{t+1}g_{t+1}=\rho$
	\end{algorithmic}
\end{algorithm}

\begin{algorithm}
	\caption{Limited-BFGS  \cite{liu1989limited}} \label{L-BFGS}
	\begin{algorithmic}
		\Require ~~
		$\theta_0\in R^n$, $\epsilon>0$
	
		\Ensure ~~
		the solution $\theta^*$\\
	    $t=0$\\
		$g_0=\nabla f_0$\\
        $u_0 = \mathbf{1}$\\
        $s_0 = \mathbf{1}$
		\While {$\parallel g_t\parallel<\epsilon$}
		\State Choose $H_t^0$, for example $H_t^0=\frac{s_t^{\top} u_t}{\parallel u_t \parallel^2}I$
		\State $g_t=\nabla f_t$
		\State $d_t=-H_tg_t$ from Algorithm L-BFGS two-loop recursion for $H_tg_t$
		\State Search a step size $\eta_t$ through Wolfe rule
		\State $\theta_{t+1}=\theta_t+\eta_td_t$
		\If{$k>p$}
		\State Discard the vector pair $\{s_{t-p}, y_{t-p}\}$ from storage
		
		\EndIf
		\State Compute and save
		\State $s_t=\theta_{t+1}-\theta_t,u_t=g_{t+1}-g_t$
		\State $t=t+1$
		\EndWhile
		
	\end{algorithmic}
\end{algorithm}
\par
For more information about BFGS and L-BFGS algorithms, one can refer to \cite{nocedal2006numerical,nocedal1980updating}. Recently, the batch L-BFGS on machine learning was proposed \cite{berahas2016multi}, which uses the overlapping mini-batches for consecutive samples for quasi-Newton update. It means that the calculation of $u_t$ becomes
$
u_t=\nabla_{S_{t+1}}f(\theta_{t+1})-\nabla_{S_t}f(\theta_t),
$
where $S_t$ is a small subset of samples, meanwhile $S_{t+1}$ and $S_t$ are not independent, perhaps containing a relatively large overlap.
Some numerical results in \cite{berahas2016multi} have shown that the modification in L-BFGS is effective in practice.

\subsubsection{Stochastic Quasi-Newton Method}
\par

In many large-scale machine learning models, it is necessary to use a stochastic approximation algorithm with each step of update based on a relatively small training subset \cite{byrd2016stochastic}. Stochastic algorithms often obtain the best generalization performances in large-scale learning systems \cite{bottou2008tradeoffs}. The quasi-Newton method only uses the first-order gradient information to approximate the Hessian matrix. It is a natural idea to combine the quasi-Newton method with the stochastic method, so that it can perform on large-scale problems. Online-BFGS and online-LBFGS are two variants of BFGS \cite{schraudolph2007stochastic}.
\par
Consider the minimization of a convex stochastic function,
\begin{equation}
\mathrm{min}_{\theta \in \mathbb{R}} \ \ F(\theta) =\mathbb{E}[f(\theta,\xi)],
\end{equation}
where $\xi$ is a random seed. We assume that $\xi$ represents a sample (or a set of samples) consisting of an input-output pair $(x,y)$. In machine learning $x$ typically represents an input and $y$ is the target output. $f$ usually has the following form:
\begin{equation}
f(\theta;\xi)=f(\theta;x_i,y_i)=l(h(w;x_i);y_i),
\end{equation}
where $h$ is a prediction model parameterized by $\theta$, and $l$ is a loss function. We define $f_i(\theta)=f(\theta;x_i,y_i)$, and use the empirical loss to define the objective,

\begin{equation}
F(\theta)=\frac{1}{N}\sum_{i=1}^N f_i(\theta).
\end{equation}

Typically, if a large amount of training data is used to train the machine learning models, a better choice is using mini-batch stochastic gradient,
\begin{equation}
\nabla F_{S_t}(\theta_t)=\frac{1}{c}\sum_{i\in S_t}\nabla f_i(\theta_t),
\end{equation}
where subset $S_t\subset \{1,2,3 \cdots N\}$ is randomly selected. $c$ is the cardinality of $S_t$ and $c \ll N$. Let $S_t^H\subset \{1,2,3,\cdots,N\}$ be a randomly chosen subset of the training samples and the stochastic Hessian estimate can be
\begin{equation}
\nabla^2 F_{S_t}(\theta_t)=\frac{1}{c_h}\sum_{i\in S_t^H}\nabla^2 f_i(\theta_t),
\end{equation}
where $c_h$ is the cardinality of $S_t^H$.
With given stochastic gradient,
a direct approach to develop stochastic quasi-Newton method is to transform deterministic gradients into stochastic gradients throughout the iterations, such as online-BFGS and online-LBFGS \cite{schraudolph2007stochastic}, which are two stochastic adaptations of the BFGS algorithms. Specifically, following the BFGS described in the previous section, $s_t,u_t$ are modified as
\begin{equation}
s_t:=\theta_{t+1}-\theta_t \quad  \mathrm{and}\quad u_t:=\nabla F_{S_t}(\theta_{t+1})-\nabla F_{S_t}(\theta_t).
\end{equation}

\par
\par
One disadvantage of this method is that each iteration requires two gradient estimates. Besides this, a more worrying fact is that updating the inverse Hessian approximations in each step may not be reasonable \cite{bottou2016optimization}. Then the stochastic quasi-Newton (SQN) method was proposed, which is to use sub-sampled Hessian-vector products to update $H_t$ by the LBFGS according to \cite{byrd2016stochastic}. Meanwhile, the authors proposed an effective approach that decouples the stochastic gradient and curvature estimate calculations to obtain a stable Hessian approximation. In particular, since

\begin{equation}
\nabla F(\theta_{t+1})-\nabla F(\theta_t)\approx
\nabla^2 F(\theta_t)(\theta_{t+1}-\theta_t),
\end{equation}
$u_t$ can be rewritten as
\begin{equation}
u_t:=\nabla^2 F_{S_t^H}(\theta_t)s_t.
\end{equation}
Based on these techniques, an SQN Framework was proposed, and the detailed procedure is shown in Algorithm \ref{SQN}.
\begin{algorithm}
	\caption{SQN Framework \cite{bottou2016optimization}}  \label{SQN}
	\begin{algorithmic}
		\Require ~~ $\theta_0$, $V$, $m$, ${\eta_t}$
		\Ensure ~~	The solution $\theta^*$
		\For{t=1, 2, 3, 4,.....,}
		\State $s_t'=H_tg_t$ using the two-loop recursion.
		\State $s_t=-\eta_t s_t'$
		\State $\theta_{t+1}=\theta_t+s_t'$
		\If {update pairs}
		\State Compute $s_t$ and $u_t$
		\State Add a new displacement pair $\{s_t,u_t\}$ to $V$
		\If{$|V|>m$}
		\State Remove the eldest pair from $V$
		\EndIf
		\EndIf
		\EndFor
	\end{algorithmic}
\end{algorithm}
\par


In the above algorithm, $V=\{s_t,u_t\}$ is a collection of $m$ displacement pairs, and $g_t$ is the current stochastic gradient $\nabla F_{S_t}(\theta_t)$. Meanwhile, the matrix-vector product $H_tg_t$ can be computed by a two-loop recursion as described in the previous section. Recently, more and more work has achieved very good results in stochastic quasi-Newton. Specifically, a regularized stochastic
BFGS method was proposed, which makes a corresponding analysis of the convergence of this optimization method \cite{mokhtari2014res}. Further, an online L-BFGS was presented in \cite{mokhtari2015global}. A linearly convergent method was proposed \cite{moritz2016linearly}, which combines the L-BFGS method in \cite{byrd2016stochastic} with the variance reduction technique. Besides these, a variance reduced block L-BFGS method was proposed, which works by employing the actions of a sub-sampled Hessian on a set of random vectors \cite{gower2016stochastic}.
\par
To sum up, we have discussed the techniques of using stochastic methods in second-order optimization. The stochastic quasi-Newton method is a combination of the stochastic method and the quasi-Newton method, which makes the quasi-Newton method extend to large datasets. We have introduced the related work of the stochastic quasi-Newton method in recent years, which reflects the potential of the stochastic quasi-Newton method in machine learning applications.

\subsubsection{Hessian-Free Optimization Method}
The main idea of Hessian-free (HF) method is similar to Newton's method, which employs second-order gradient information. The difference is that the HF method is not necessary to directly calculate the Hessian matrix $H$. It estimates the product $Hv$ by some techniques, and thus is called ``Hessian free''.

Consider a local quadratic approximation $Q_{\theta}(d_t)$ of the object $F$ around parameter $\theta$,
	
	\begin{equation}
	F(\theta_t+d_t)\approx Q_{\theta}(d_t)=F(\theta_t)+\nabla F(\theta_t)^{\top}  d_t+\frac{1}{2}d_t^{\top}  B_{t}d_t,
	\end{equation}
where $d_t$ is the search direction.
The HF method applies the conjugate gradient method to compute an approximate solution $d_t$ of the linear system,
\begin{equation}
B_td_t=-\nabla F(\theta_t),\label{3.2.4}
\end{equation}
where $B_t=H(\theta_t)$ is the Hessian matrix, but in practice $B_t$ is often defined as $B_t=H(\theta_t)+\lambda I,\ \lambda\geq 0$ \cite{martens2010deep}.
The new update is then given by
\begin{equation}
\theta_{t+1}=\theta_{t}+\eta_t d_t,
\end{equation}
where $\eta_t$ is the step size that ensures
sufficient decrease in the objective function, usually obtained by a linear search. According to \cite{martens2010deep}, the basic framework of HF optimization is shown in Algorithm \ref{HF-method}.

\begin{algorithm}
	\caption{Hessian-Free Optimization Method \cite{martens2010deep}}\label{HF-method}
	\begin{algorithmic}
		\Require~~$\theta_0$, $\nabla f(\theta_0)$, $\lambda$

		\Ensure~~The solution $\theta^*$
	    \State $t=0$
		\Repeat
		\State $g_t=\nabla f(\theta_t)$
		\State Compute $\lambda$ by some methods
		\State $B_t(v)\equiv H(\theta_t)v+\lambda v $
        \State Compute the step size $\eta_t$
		\State $d_t=CG(B_t,-g_t)$
		\State $\theta_{t+1}=\theta_t+\eta_t d_t$
		\State $t=t+1$
		\Until satisfy convergence condition
	\end{algorithmic}
\end{algorithm}
\par
The advantage of using the conjugate gradient method is that it can calculate the Hessian-vector product without directly calculating the Hessian matrix. Because in the CG-algorithm, the Hessian matrix is paired with a vector, then we can compute the Hessian-vector product to avoid the calculation of the Hessian inverse matrix. There are many ways to calculate Hessian-vector products, one of which is calculated by a finite difference as \cite{martens2010deep}

\begin{equation}
Hv={\lim_{\varepsilon \to +0}}\frac{\nabla f(\theta+\varepsilon v)-\nabla f(\theta)}{\varepsilon}.
\end{equation}
\par
\textbf{Sub-sampled Hessian-Free Method}
HF is a well-known method, and has been studied for decades in the optimization literatures, but has shortcomings when applied to deep neural networks with large-scale data \cite{martens2010deep}. Therefore, a sub-sampled technique is employed in HF, resulting in an efficient HF method \cite{martens2010deep,byrd2011use}. The cost in each iteration can be reduced by using only a small sample set $S$ to calculate $Hv$.
The objective function has the following form:
\begin{equation}
\mathrm{min} \ \ F(\theta)=\frac{1}{N}\sum ^N_{i=1}f_i(\theta).
\end{equation}
In the $t$th iteration, the stochastic gradient estimation can be written as
\begin{equation}
\nabla F_{S_t}(\theta_t)=\frac{1}{|S_t|}\sum_{i\in S_t}\nabla f_i(\theta_t),
\end{equation}
and the stochastic Hessian estimate is expressed as
\begin{equation}
\nabla^2 F_{S^H_t}(\theta_t)=\frac{1}{|S^H_t|}\sum_{i\in S^H_t}\nabla^2 f_i(\theta_t).
\end{equation}
As described above, we can obtain the approximate solution of direction $d_t$ by employing the CG method to solve the linear system,
\begin{equation}
\nabla^2 F_{S^H_t}(\theta_t)d_t=-\nabla F_{S_t}(\theta_t),
\end{equation}
in which the stochastic gradient and stochastic Hessian matrix are used. The basic framework of sub-sampled HF algorithm is given in \cite{byrd2011use}.
\par
A natural question is how to determine the size of $S^H_t$. On one hand, $S^H_t$ can be chosen small enough so that the total cost of CG iteration is not much greater than a gradient evaluation. On the other hand, $S^H_t$ should be large enough to get useful curvature information from Hessian-vector product. How to balance the size of $S^H_t$ is a challenge being studied \cite{byrd2011use}.
\subsubsection{Natural Gradient}

The natural gradient method can be potentially applied to any objective function which measures the performance of some statistical models \cite{amari1998natural}. It enjoys richer theoretical properties when applied to objective functions based on the KL divergence between the model's distribution and the target distribution, or certain approximation surrogates of these \cite{martens2014new}.

The traditional gradient descent algorithm is based on the Euclidean space. However, in many cases, the parameter space is not Euclidean, and it may have a Riemannian metric structure. In this case, the steepest direction of the objective function cannot be given by the ordinary gradient and should be given by the natural gradient \cite{amari1998natural}.


\par
We consider such a model distribution $p(y|x,\theta)$, and $\pi(x,y)$ is an empirical distribution. We need to fit the parameters $\theta \in R^N $. Assume that $x$ is an observation vector, and $y$ is its associated label. It has the objective function,
\begin{equation}
F(\theta)=\mathbb{E}_{(x,y)\sim \pi}[-\log p(y|x,\theta)],
\end{equation}
and we need to solve the optimization problem,
\begin{equation}
\theta^* =  \mathrm{argmin}_\theta F(\theta).
\end{equation}
According to \cite{amari1998natural}, the natural gradient can be transformed from a traditional gradient multiplied by a Fisher information matrix, i.e.,
\begin{equation}
\nabla_N F=G^{-1}\nabla F,
\end{equation}
where $F$ is the object function, $\bigtriangledown F$ is the traditional gradient, $\bigtriangledown_N F$ is the natural gradient, and $G$ is the Fisher information matrix, with the following form:
\begin{equation}
G=\mathbb{E}_{x\sim \pi}\left[{\mathbb{E}_{y\sim p(y|x,\theta)}\left[(\frac{\partial p(y|x;\theta)}{\partial \theta})(\frac{\partial p(y|x;\theta)}{\partial \theta})^{\top}\right]}\right].
\end{equation}

The update formula with the natural gradient is
\begin{equation}
\theta_t=\theta_t-\eta_t \nabla_N F.
\end{equation}
We cannot ignore that the application of the natural gradient is very limited because of too much computation. It is expensive to estimate the Fisher information matrix and calculate its inverse matrix. To overcome this limitation, the truncated Newton's method was developed \cite{martens2010deep}, in which the inverse is calculated by an iterative procedure, thus avoiding the direct calculation of the inverse of the Fisher information matrix. In addition, the factorized natural gradient (FNG) \cite{grosse2015scaling} and Kronecker-factored approximate curvature
(K-FAC) \cite{martens2015optimizing} methods were proposed to use the derivatives of
probabilistic models to
calculate the approximate natural gradient update.

\subsubsection{Trust Region Method}
The update process of most methods introduced above can be described as $\theta_t+\eta_t d_t$.
The displacement of the point in the direction of $d_t$ can be written as $s_t$.
The typical trust region method (TRM) can be used for unconstrained nonlinear optimization problems \cite{sun2006optimization, byrd2000trust, hei2007practical}, in which the displacement $s_t$ is directly determined without the search direction $d_t$.
\par

For the problem $\min f_{\theta}(x)$, the TRM \cite{sun2006optimization} uses the second-order Taylor expansion to approximate the objective function $f_{\theta}(x)$, denoted as $q_{t}(s)$. Each search is done within the range of trust region with radius $\bigtriangleup_t$. This problem can be described as

\begin{equation}
\left\{
\begin{aligned}
\min \quad  q_{t}(s) &=f_{\theta}(x_t)+g_t^\top s+\frac{1}{2}s^\top B_ts \label{3.4.1},\\
s.t. \ \ \ \quad ||s_t|| &\leq \bigtriangleup _t,
\end{aligned}
\right.
\end{equation}
\noindent
where $g_t$ is the approximate gradient of the objective function $f(x)$ at the current iteration point $x_t$, $g_t \approx \nabla f(x_t)$, $B_t$ is a symmetric matrix, which is the approximation of Hessian matrix $\nabla^2 f_{\theta}(x_t)$, and $\bigtriangleup _t > 0$ is the radius of the trust region. If the $L_2$ norm is used in the constraint function, it becomes the Levenberg-Marquardt algorithm \cite{lourakis2005brief}.

\par
If $s_t$ is the solution of the trust region subproblem (\ref{3.4.1}),  the displacement $s_t$ of each update is limited by the trust region radius $\bigtriangleup _t$. The core part of the TRM is the update of $ \bigtriangleup _t$. In each update process, the similarity of the quadratic model $q(s_t)$ and the objective function $f_{\theta}(x)$ is measured, and $ \bigtriangleup _t$ is updated dynamically.
The actual amount of descent in the $t$th iteration is \cite{sun2006optimization}
\begin{equation}
\bigtriangleup f_t=f_t-f(x_t+s_t).
\end{equation}
The predicted drop in the $t$th iteration is
\begin{equation}
\bigtriangleup q_t=f_t-q(s_t).
\end{equation}
The ratio $r_t$ is defined to measure the approximate degree of both,
\begin{equation}
r_t=\frac{\bigtriangleup f_t}{\bigtriangleup q_t}.
\end{equation}
It indicates that the model is more realistic than expected when $r_t$ is close to 1, and then we should consider expanding $ \bigtriangleup _t$. At the same time, it indicates that the model predicts a large drop and the actual drop is small when $r_t$ is close to 0, and then we should reduce $ \bigtriangleup _t$. Moreover, if $r_t$ is between 0 and 1, we can leave $ \bigtriangleup _t$ unchanged. The thresholds 0 and 1 are generally set as the left and right boundaries of $r_t$ \cite{sun2006optimization}.
\subsubsection{Summary}
We summarize the mentioned high-order optimization methods in terms of properties, advantages and disadvantages in Table \ref{3.2.8}.
\begin{table*}

			\caption{Summary of High-Order Optimization Methods }	\label{3.2.8}
		
	\begin{center}
		\begin{tabular}{p{2cm}p{4.5cm}p{4.5cm}p{4.5cm}}

			 \toprule
			Method&Properties
&            Advantages& Disadvantages  \\
			\midrule
		
			& & & \\

		   {Conjugate Gradient \cite{hestenes1952methods}}
			& It is an optimization method between the first-order and second-order gradient methods. It constructs a set of conjugated directions using the gradient of known points, and searches along the conjugated direction to find the minimum points of the objective function.
		   &CG method only calculates the first order gradient but has faster convergence than the steepest descent method.
	 	   &Compared with the first-order gradient method, the calculation of the conjugate gradient is more complex.
			\\
			\hline
			& & & \\
			Newton's Method \cite{avriel2003nonlinear}
			& Newton's method calculates the inverse matrix of the Hessian matrix to obtain faster convergence than the first-order gradient descent method.
			&Newton's method uses second-order gradient information which has faster convergence than the first-order gradient method. Newton's method has quadratic convergence  under certain conditions.
			& It needs long computing time and large storage space to calculate and store the inverse matrix of the Hessian matrix at each iteration.
		\\
			\hline
			& & & \\
			Quasi-Newton Method \cite{nocedal2006numerical}
			&Quasi-Newton method uses an approximate matrix to approximate the the Hessian matrix or its inverse matrix. Popular quasi-Newton methods include DFP, BFGS and LBFGS.
			&Quasi-Newton method does not need to calculate the inverse matrix of the Hessian matrix, which reduces the computing time. In general cases, quasi-Newton method can achieve superlinear convergence.
			&Quasi-Newton method needs a large storage space, which is not suitable for handling the optimization of large-scale problems.
			\\
			\hline
			& & & \\
		Sochastic Quasi-Newton Method \cite{bottou2016optimization}.
			& Stochastic quasi-Newton method employs techniques of stochastic optimization. Representative methods are online-LBFGS \cite{schraudolph2007stochastic} and SQN \cite{byrd2016stochastic}.
			&Stochastic quasi-Newton method can deal with large-scale machine learning problems.
			& Compared with the stochastic gradient method, the calculation of stochastic quasi-Newton method is more complex.
			\\
			\hline
			& & &\\
			Hessian Free Method \cite{martens2010deep}
			&HF method performs a sub-optimization using the conjugate gradient, which avoids the expensive computation of inverse Hessian matrix.
			&HF method can employ the second-order gradient information but does not need to directly calculate Hessian matrices. Thus, it is suitable for high dimensional optimization.
			&
				The cost of computation for the matrix-vector product in HF method increases linearly with the increase of training data. It does not work well for large-scale problems. \\
			\hline
				& & &\\
			Sub-sampled Hessian Free Method \cite{byrd2011use}
			&
				Sup-sampled Hessian free method uses stochastic gradient and sub-sampled Hessian-vector during the process of updating.
			&The sub-sampled HF method can deal with large-scale machine learning optimization problems.
			&Compared with the stochastic gradient method, the calculation is more complex and needs more computing time in each iteration. \\
			\hline
			
			& & & \\
			Natural Gradient \cite{amari1998natural}
			&The basic idea of the natural gradient is to construct the gradient descent algorithm in the predictive function space rather than the parametric space.
			&The natural gradient uses the Riemann structure of the parametric space to adjust the update direction, which is more suitable for finding the extremum of the objective function.
			&In the natural gradient method, the calculation of the Fisher information matrix is complex.
		\\
			\hline
		\end{tabular}
	\end{center}
	\vspace{-0.5cm}

\end{table*}

\subsection{Derivative-Free Optimization}

\par
For some optimization problems in practical applications, the derivative of the objective function may not exist or is not easy to calculate. The solution of finding the optimal point, in this case, is called derivative-free optimization, which is a discipline of mathematical optimization \cite{conn2009introduction, audet2016blackbox, Rios2013}. It can find the optimal solution without the gradient information.
\par
There are mainly two types of ideas for derivative-free optimization.
One is to use heuristic algorithms. It is characterized by empirical rules and chooses methods that have already worked well, rather than derives solutions systematically. There are many types of heuristic optimization methods, including classical simulated annealing arithmetic, genetic algorithms, ant colony algorithms, and particle swarm optimization \cite{kirkpatrick1983optimization,mitchell1998introduction,dorigo2008ant}. These heuristic methods usually yield approximate global optimal values, and theoretical support is weak. We do not focus on such techniques in this section.
The other is to fit an appropriate function according to the samples of the objective function.
This type of method usually attaches some constraints to the search space to derive the samples.
Coordinate descent method is a typical derivative-free algorithm \cite{bertsekas1999nonlinear}, and it can be extended and applied to optimization algorithms for machine learning problems easily. In this section, we mainly introduce the coordinate descent method.

\par

The coordinate descent method is a derivative-free optimization algorithm for multi-variable functions. Its idea is that a one-dimensional search can be performed sequentially along each axis direction to obtain updated values for each dimension. This method is suitable for some problems in which the loss function is non-differentiable.
\par

The vanilla approach is to select a set of bases $e_1, e_2, ..., e_D$ in the linear space as the search directions and minimizes the value of the objective function in each direction. For the target function $L(\Theta)$, when $\Theta^t$ is already obtained, the $j$th dimension of $\Theta^{t+1}$ is solved by \cite{conn2009introduction}
\begin{equation}
\theta_j^{t+1}={\arg \min}_{\theta_{j}\in R} L(\theta_1^{t+1},...,\theta_{j-1}^{t+1}, \theta_{j},\theta_{j+1}^{t},...,\theta_{D}^{t}).
\end{equation}
Thus, $L(\Theta^{t+1}) \leq L(\Theta^{t}) \leq ... \leq L(\Theta^{0}) $ is guaranteed. The convergence of this method is similar to the gradient descent method.
The order of update can be an arbitrary arrangement from $e_1$ to $e_D$ in each iteration. The descent direction can be generalized from the coordinate axis to the coordinate block \cite{richtarik2014iteration}.

\par

The main difference between the coordinate descent and the gradient descent is that each update direction in the gradient descent method is determined by the gradient of the current position, which may not be parallel to any coordinate axis. In the coordinate descent method, the optimization direction is fixed from beginning to end. It does not need to calculate the gradient of the objective function. In each iteration, the update is only executed along the direction of one axis, and thus the calculation of the coordinate descent method is simple even for some complicated problems. For indivisible functions, the algorithm may not be able to find the optimal solution in a small number of iteration steps.
An appropriate coordinate system can be used to accelerate the convergence. For example, the adaptive coordinate descent method takes principal component analysis to obtain a new coordinate system with as little correlation as possible between the coordinates \cite{loshchilov2011adaptive}. The coordinate descent method still has limitations when performing on the non-smooth objective function, which may fall into a non-stationary point.

\subsection{Preconditioning in Optimization}
Preconditioning is a very important technique in optimization methods. Reasonable preconditioning can reduce the iteration number of optimization algorithms. For many important iterative methods, the convergence depends largely on the spectral properties of the coefficient matrix \cite{huckle1999approximate}. It can be simply considered that the pretreatment is to transform a difficult linear system $A\theta=b$ into an equivalent system with the same solution but better spectral characteristics. For example, if $M$ is a nonsingular approximation of the coefficient matrix $A$, the transformed system,
\begin{equation}
M^{-1}A\theta=M^{-1}b,\label{pre-con}
\end{equation}
will have the same solution as the system $A\theta=b$. But (\ref{pre-con}) may be easier to solve and the spectral properties of the coefficient matrix $M^{-1}A$ may be more favorable.

In most linear systems, e.g., $A\theta=b$, the matrix $A$ is often complex and makes it hard to solve the system. Therefore, some transformation is needed to simplify this system. $M$ is called the preconditioner. If the matrix after using preconditioner is obviously structured, or sparse, it will be beneficial to the calculation \cite{benzi2002preconditioning}.

The conjugate gradient algorithm mentioned previously is the most commonly used optimization method with preconditioning technology, which speeds up the convergence. The algorithm is shown in Algorithm \ref{PCG}.

\begin{algorithm}
	\caption{Preconditioned Conjugate Gradient Method \cite{nocedal2006numerical}} \label{PCG}
	\begin{algorithmic}
		\Require ~~$A$, $\theta_0$, $M$, $b$
		\Ensure~~The solution $\theta^*$
\State $f_0=f(\theta_0)$
\State $g_0=\nabla f(\theta_0)=A\theta_0-b$
		\State $y_0$ is the solution of $My=g_0$
		\State $d_0=-g_0$
		\State $t=t$
		\While {$g_t\neq 0$}
		\State  $\eta_t=\frac{g_t^{\top} y_t}{d_t^{\top} Ad_t}$
		\State $\theta_{t+1}=\theta_t+\eta_td_t$
		\State $g_{t+1}=g_t+\eta_tAd_t$
		\State $y_{t+1}=$solution of $My=g_t$
		\State $\beta_{t+1}=\frac{g_{t+1}^{\top} y_{t+1}}{g_{t}^{\top}  d_{t}}$
		\State $d_{t+1}=-y_{t+1}+\beta_{t+1}d_t$
		\State $t=t+1$
		\EndWhile

	\end{algorithmic}
\vspace{0.1cm}
\end{algorithm}

\subsection{Public Toolkits for Optimization}

Fundamental optimization methods are applied in machine learning problems extensively. There are many integrated powerful toolkits. We summarize the existing common optimization toolkits and present them in Table \ref{3.5}.
\begin{table*}
	
	\caption{Available Toolkits for Optimization }
	\begin{center}
		\begin{tabular}{lp{3cm}p{7cm}p{7cm}}
			\toprule
			Toolkit& Language& Description\\
			\midrule
			\label{3.5}
			& & \\
			CVX \cite{grant2014cvx} &Matlab&CVX is a matlab-based modeling system  for convex optimization but cannot handle large-scale problems.\\ & &
			\url{http://cvxr.com/cvx/download/}\\ \hline
			& & \\
			CVXPY \cite{diamond2016cvxpy}&Python&CVXPY is a python package developed by Stanford University Convex Optimization Group for solving convex optimization problems.\\ & &
			\url{http://www.cvxpy.org/}\\ \hline
			& & \\
			CVXOPT \cite{andersen2013cvxopt}&Python& CVXOPT can be used for handling convex optimization. It is developed by Martin Andersen, Joachim Dahl, and Lieven Vandenberghe. \\ & &
			\url{http://cvxopt.org/}\\ \hline
			& & \\
			APM \cite{hedengren2014nonlinear}&Python&APM python is suitable for large-scale optimization and can solve the problems of linear programming, quadratic programming, integer programming, nonlinear optimization and so on.\\ & &
			\url{http://apmonitor.com/wiki/index.php/Main/PythonApp}\\ \hline
			& & \\
			SPAMS \cite{mairal2014spams}&
			C++&SPAMS is an optimization toolbox for solving various sparse estimation problems, which is developed and maintained by Julien Mairal. Available interfaces include matlab, R, python and C++.\\ & &
			\url{http://spams-devel.gforge.inria.fr/}\\ \hline
			& & \\
			minConf&Matlab&minConf can be used for optimizing differentiable multivariate functions which subject to simple constraints on parameters. It is a set of matlab functions, in which there are many methods to choose from.\\ & &
			\url{https://www.cs.ubc.ca/~schmidtm/Software/minConf.html}\\ \hline
			& & \\
			tf.train.optimizer \cite{abadi2016tensorflow}&Python; C++; CUDA&The basic optimization class, which is usually not called directly and its subclasses are often used. It includes classic optimization algorithms such as gradient descent and AdaGrad.\\ & & \url{https://www.tensorflow.org/api_guides/python/train}\\
			\hline
		\end{tabular}
	\end{center}
\vspace{0.5cm}
\end{table*}

\section{Developments and Applications for Selected Machine Learning Fields}\label{sec:development}

Optimization is one of the cores of machine learning. Many optimization methods are further developed in the face of different machine learning problems and specific application environments. The machine learning fields selected in this section mainly include deep neural networks, reinforcement learning, variational inference and Markov chain Monte Carlo.

\subsection{Optimization in Deep Neural Networks}
The deep neural network (DNN) is a hot topic in the machine learning community in recent years. There are many optimization methods for DNNs. In this section, we introduce them from two aspects, i.e., first-order optimization methods and high-order optimization methods.
\subsubsection{First-Order Gradient Method in Deep Neural Networks}
The stochastic gradient optimization method and its adaptive variants have been widely used in DNNs and have achieved good performance. SGD introduces the learning rate decay factor and AdaGrad accumulates all previous gradients so that their learning rates are continuously decreased and converge to zero. However, the learning rates of these two methods make the update slow in the later stage of optimization. AdaDelta, RMSProp, Adam and other methods use the exponential averaging to provide effective updates and simplify the calculation. These methods use exponential moving average to alleviate the problems caused by the rapid decay of the learning rate but limit the current learning rate to only relying on a few gradients \cite{reddi2018convergence}. Reddi et al. used a simple convex optimization example to demonstrate that the RMSProp and Adam algorithms could not converge \cite{reddi2018convergence}. Almost all the algorithms that rely on a fixed-size window of the past gradients will suffer from this problem, including AdaDelta and Nesterov-accelerated adaptive moment estimation (Nadam) \cite{dozat2016incorporating}.

It is better to rely on the long-term memory of past gradients rather than the exponential moving average of gradients to ensure convergence. A new version of Adam \cite{reddi2018convergence}, called AmsGrad, uses a simple correction method to ensure the convergence of the model while preserving the original computational performance and advantages. Compared with the Adam method, the AmsGrad makes the following changes to the first-order moment estimation and the second-order moment estimation:

\begin{equation}
\left\{
\begin{aligned}
m_t &=\beta_{1t} m_{t-1}+(1-\beta_{1t} )g_t, \\
V_t &=\sqrt{ \beta_2V_{t-1}+(1-\beta_2 )g_t^2}, \\
\hat{V}_t &=\max (\hat{V}_{t-1},V_t),
\end{aligned}
\right.
\end{equation}
where $\beta_{1t}$ is a non-constant which decreases with time, and $\beta_{2}$ is a constant learning rate. The correction is operated in the second-order moment $V_t$, making $\hat{V}_t$ monotonous. $\hat{V}_t$ is substantially used in the iteration of the target function. The AmsGrad method takes the long-term memory of past gradients based on the Adam method, guarantees the convergence in the later stage, and works well in applications.

\par
Further, adjusting parameters $\beta_1,\beta_2$ at the same time helps to converge to a certain extent. For example, $\beta_1$ can decay modestly as $\beta_{1t}=\frac{\beta_1 }{t} $, $\beta_{1t} \leq \beta_1 $, for all $t \in [T]$. $\beta_2$ can be set as $\beta_{2t}=1-\frac{1}{t}$, for all $t\in [T] $, as in AdamNC algorithm \cite{reddi2018convergence}.
\par
Another idea of combining SGD and Adam was proposed for solving the non-convergence problem of adaptive gradient algorithm \cite{keskar2017improving}. Adaptive algorithms, such as Adam, converge fast and are suitable for processing sparse data. SGD with momentum can converge to more accurate results. The combination of SGD and Adam develops the advantages of both methods. Specifically, it first trains with Adam to quickly drop and then switches to SGD for precise optimization based on the previous parameters at an appropriate switch point. The strategy is named as switching from Adam to SGD (SWATS) \cite{keskar2017improving}.
There are two core problems in SWATS. One is when to switch from Adam to SGD, the other is how to adjust the learning rate after switching the optimization algorithm. The SWATS approach is described in detail below.
\par
The movement $d^{Adam}$ of the parameter at iteration $t$ of the Adam is
\begin{equation}
d_t^{Adam}=\frac{\eta^{Adam}}{{V_t}} m_t,
\end{equation}
where $\eta^{Adam}$ is the learning rate of Adam \cite{keskar2017improving}.
The movement $d^{SGD}$ of the parameter at iteration $t$ of the SGD is
\begin{equation}
d_t^{SGD}=\eta^{SGD} g_t,
\end{equation}
where $\eta^{SGD}$ is the learning rate of SGD and $g_t$ is the gradient of the current position \cite{keskar2017improving}.
\par
The movement of SGD can be decomposed into the learning rates along Adam's direction and its orthogonal direction. If SGD is going to finish the trajectory but Adam has not finished due to the momentum after selecting the optimization direction, walking along Adam's direction is a good choice for SWATS. At the same time, SWATS also adjusts its optimized trajectory by moving in the orthogonal direction. Let
\begin{equation}
Proj_{Adam} \ d_t^{SGD}=d_t^{Adam},
\end{equation}
and derive solution
\begin{equation}
\eta_t^{SGD}=\frac{(d_t^{Adam})^\mathrm{T} d_t^{Adam}}{(d_t^{Adam})^\mathrm{T} g_t},
\end{equation}
where $Proj_{Adam}$ means the projection in the direction of Adam. To reduce noise, a moving average can be used to correct the estimate of the learning rate,

\begin{equation}
\lambda_t^{SGD}=\beta_2 \lambda_{t-1}^{SGD}+(1-\beta_2)\eta_t^{SGD},
\end{equation}

\begin{equation}
\tilde{\lambda_t}^{SGD}=\frac{\lambda_t^{SGD}}{1-\beta_2},
\end{equation}	
where $\lambda_t^{SGD}$ is the first moment of learning rate $\eta^{SGD}$, and $\tilde{\lambda_t}^{SGD}$ is the learning rate of SGD after converting \cite{keskar2017improving}. For switch point, a simple guideline $|\tilde{\lambda_t}^{SGD}-\lambda_t^{SGD}|< \epsilon$ is often used \cite{keskar2017improving}. Although there is no rigorous mathematical proof for selecting this conversion criterion, it performs well across a variety of applications. For the mathematical proof of switch point, further research can be conducted. Although the SWATS is based on Adam, this switching method is also applicable to other adaptive methods, such as AdaGrad and RMSProp.
The procedure is insensitive to hyper-parameters and can obtain an optimal solution comparable to SGD, but with a faster training speed in the case of deep networks.
\par
Recently some researchers are trying to explain and improve the adaptive methods \cite{loshchilov2017fixing,zhang2017normalized}. Their strategies can also be combined with the above switching techniques to enhance the performance of the algorithm.

General fully connected neural networks cannot process sequential data such as text and audio. {Recurrent neural network (RNN)} is a kind of neural networks that is more suitable for processing sequential data. It was generally considered that the use of first-order methods to optimize RNN was not effective, because the SGD and its variant methods were difficult to learn long-term dependencies in sequence problems \cite{ sutskever2013training,sutskever2013importance, salehinejad2017recent}.

\par

 In recent years, a well-designed method of random parameter initialization scheme using only SGD with momentum without curvature information has achieved good results in training RNNs \cite{sutskever2013training}. In \cite{sutskever2013importance,bengio2013advances}, some techniques for improving the optimization in training RNNs are summarized such as the momentum methods and NAG.
The first-order optimization methods have got development for training RNNs, but they still face the problem of slow convergence in deep RNNs. The high-order optimization methods employing curvature information can accelerate the convergence near the optimal value and is considered to be more effective in optimizing DNNs.

\subsubsection{High-Order Gradient Method in Deep Neural Networks}

We have described the first-order optimization method applied in DNNs. As most DNNs use large-scale data, different versions of stochastic gradient methods were developed and have got excellent performance and properties. For making full use of gradient information, the second-order method is gradually applied to DNNs. In this section, we mainly introduce the Hessian-free method in DNN.

Hessian-free (HF) method has been studied for a long time in the field of optimization, but it is not directly suitable for dealing with neural networks \cite{martens2010deep}. As the objective function in DNN is not convex, the exact Hessian matrix may not be positive definite. Therefore, some modifications need to be made so that the HF method can be applied to neural networks \cite{martens2012training}.
\par
\textbf{The Generalized Gauss-Newton Matrix}
One solution is to use the generalized Gauss-Newton (GGN) matrix, which can be seen as an approximation of a Hessian matrix \cite{schraudolph2002fast}. The GGN matrix is a provably positive semidefinite matrix, which avoids the trouble of negative curvature.  There are at least two ways to derive the GGN matrix \cite{martens2012training}. Both of them require that $f(\theta)$ can be expressed as a composition of two functions written as $f(\theta)=Q(F(\theta))$ where $f(\theta)$ is the object function and $Q$ is convex. The GGN matrix $G$ takes the following form,
\begin{equation}
G=J^{\top}  Q''J,
\end{equation}
where $J$ is the Jacobian of $F$.
\par
\textbf{Damping Methods}
Another modification to the HF method is to use different damping methods. For example, Tikhonov damping, one of the most famous damping methods, is implemented by introducing a quadratic penalty term into the quadratic model. A quadratic penalty term $\frac{\lambda}{2} d^{\top}d$ is added to the quadratic model,

\begin{equation}
Q(\theta):=Q(\theta)+\frac{\lambda}{2}d^{\top}  d=f(\theta_t)+\nabla f(\theta_t)^{\top}  d+\frac{1}{2}d^{\top}  Bd,
\end{equation}
where $B=H+\lambda I$, and $\lambda>0$ determines the ``strength'' of the damping which is a scalar parameter. Thus, $Bv$ is formulated as
$Bv=(H+\lambda I)v=Hv+\lambda v$. However, the basic Tikhonov damping method is not good in training RNNs \cite{martens2011learning}.
Due to the complex structure of RNNs, the local quadratic approximation in certain directions in the parameter space, even at very small distances, maybe highly imprecise. The Tikhonov damping method can only compensate for this by increasing punishment in all directions because the method lacks a selective mechanism \cite{martens2012training}. Therefore, the structural damping was proposed, which makes the performance substantially better and more robust.
\par
The HF method with structural damping can effectively train RNNs \cite{martens2012training}.
	Now we briefly introduce the HF method with structural damping. Let $e(x,\theta)$ mean the vector-value function of $\theta$ which can be interpreted as intermediate quantities during the calculation of $f(x,\theta)$, where $f(x,\theta)$ is the object function. For instance, $e(x,\theta)$ might contain the activation function of some layers of hidden units in neural networks (like RNNs). A structural damping can be defined as
	\begin{equation}
	R(\theta)=\frac{1}{|S|}\sum_{(x,y)\in S}D(e(x,\theta),e(x,\theta_t)),
	\end{equation}
	where $D$ is a distance function or a loss function. It can prevent a large change in $e(x,\theta)$ by penalizing the distance between $e(x,\theta)$ and $e(x,\theta_t)$.
	Then, the damped local objective can be written as
	\begin{equation}
	Q_{\theta}(d)'=Q_{\theta}(d)+\mu R(d+\theta_t)+\frac{\lambda}{2}d^{\top}d,
	\end{equation}
	where $\mu$ and $\lambda$ are two parameters to be dynamically adjusted. $d$ is the direction at the $t$th iteration. More details of the structural damping can refer to \cite{martens2012training}.
	\par
	Besides, there are many second-order optimization methods employed in RNNs. For example, quasi-Newton based optimization and L-BFGS were proposed to train RNNs \cite{likas2000training,liu2018limited}.

\par
In order to make the damping method based on punishment work better, the damping parameters can be adjusted continuously. A Levenberg-Marquardt style heuristic method was used to adjust $\lambda$ directly \cite{martens2010deep}. The Levenberg-Marquardt heuristic is described as follows:
\begin{enumerate}
	\item	If $\gamma <\frac{1}{4}\lambda$ then $\lambda\gets\frac{3}{2}\lambda$,
	\item   If $\gamma >\frac{3}{4}\lambda $ then $\lambda\gets\frac{2}{3}\lambda$,
\end{enumerate}
where $\gamma$  is a ``reduction rate'' with the following form,
\begin{equation}
\gamma=\frac{f(\theta_{t-1}+d_t)-f(\theta_{t-1})}{M_{t-1}(d_t)}.
\end{equation}
\par
\textbf{Sub-sampling}
As sub-sampling Hessian can be used to handle large-scale data, several variations of the sub-sampling methods were proposed \cite{roosta2016sub,xu2016sub,bollapragada2016exact}, which used either stochastic gradients or exact gradients. These approaches use $B_t=\nabla^2_{S_t}f(\theta_t)$ as a Hessian approximation, where $S_t$ is a subset of samples.
We need to compute the Hessian-vector product in some optimization methods. If we adopt the sub-sampling method, it also means that we can save a lot of computation in each iteration, such as the method proposed in \cite{martens2010deep}.
\par
\textbf{Preconditioning}
Preconditioning can be used to simplify the optimization problems. For example, preconditioning can accelerate the CG method. It is found that diagonal matrices are particularly effective and one can use the following preconditioner \cite{martens2010deep}:
\begin{equation}
M=\big[ \mathrm{diag}(\sum^{N}_{i=1}\nabla f_i(\theta)\odot \nabla f_i(\theta))+\lambda I\big]^\alpha,
\end{equation}
where $\odot$ denotes the element-wise product and
the exponent $\alpha$ is chosen to be less than 1.

\subsection{Optimization in Reinforcement Learning}

Reinforcement learning (RL) is an important research field of machine learning and is also one of the most popular topics. Agents using deep reinforcement learning have achieved great success in learning complex behavior skills and solved challenging control tasks in high-dimensional primitive perceptual state space \cite{lillicrap2015continuous,mnih2016asynchronous,silver2016mastering}. It interacts with the environment through the trial-and-error mechanism and learns optimal strategies by maximizing cumulative rewards \cite{sutton1998reinforcement}.

\par
We describe several concepts of reinforcement learning as follows:
\begin{enumerate}[]
	\item Agent: making different actions according to the state of the external environment, and adjusting the strategy according to the reward of the external environment.
	\item Environment: all things outside the agent that will be affected by the action of the agent. It can change the state and provide the reward to the agent.
	\item State $s$: a description of the environment.
	\item Action $a$: a description of the behavior of the agent.
	\item Reward $r_t(s_{t-1},a_{t-1},s_t)$: the timely return value at time $t$.
	\item Policy $\pi (a|s)$: a function that the agent decides the action $a$ according to the current state $s$.
	\item State transition probability $p(s'|s, a)$: the probability distribution that the environment will transfer to state $s'$ at the next moment, after the agent selecting an action $a$ based on the current state $s$.

	\item $p(s',r|s,a)$: the probability that the agent transforms to state $s'$ and obtains the reward $r$, where the agent is in state $s$ and selecting the action $a$.
	
\end{enumerate}

\par
Many reinforcement learning problems can be described by Markov decision process (MDP) $<S,A,P,\gamma,r>$ \cite{sutton1998reinforcement}, in which $S$ is state space, $A$ is action space, $P$ is state transition probability function, $r$ is reward function and $\gamma$ is the discount factor $0<\gamma<1$.
At each time, the agent accepts a state and selects the action from an action set according to the policy. The agent receives feedback from the environment and then moves to the next state. The goal of reinforcement learning is to find a strategy that allows us to get the maximum $\gamma$-discounted cumulative reward. The discounted return is calculated by
\begin{equation}
G_t=\sum_{k=0}^{\infty} \gamma ^k r_{t+k}.
\end{equation}

People do not necessarily know the MDP behind the problem. From this point, reinforcement learning is divided into two categories. One is model-based reinforcement learning which knows the MDP of the whole model (including the transition probability $P$ and reward function $r$), and the other is the model-free method in which the MDP is unknown. Systematic exploration is required in the latter methods.
\par
The most commonly used value function is the state
value function,
\begin{equation}
V_{\pi}(s)=\mathbb{E}_{\pi}[G_t|S_t=s],
\end{equation}
which is the expected return of executing policy $\pi$ from state $s$.
The state-action value function is also essential which is the expected return for selecting action $a$ under state $s$ and policy $\pi$,
\begin{equation}
Q_{\pi}(s,a)=\mathbb{E}_{\pi}[G_t|S_t=s,A_t=a].
\end{equation}
The value function of the current state $s$ can be calculated by the value function of the next state $s'$. The Bellman equations of $V_{\pi}(s)$ and $Q_{\pi}(s,a)$ describe the relation by
\begin{align}
V_{\pi}(s)=&\sum_{a}\pi(a|s)\sum_{s',r}p(s',r|s,a)[r(s,a,s')\nonumber\\&+\gamma V_{\pi}(s') ],\\
Q_{\pi}(s,a)=&\sum_{s',r}p(s',r|s,a)[r(s,a,s')\nonumber\\
&+\gamma \sum_{a'}\pi(a'|s')Q_{\pi}(s',a')].
\end{align}
\par
There are many reinforcement learning methods based on value function. They are called value-based methods, which play a significant role in RL. For example, Q-learning \cite{watkins1992q} and SARSA \cite{rummery1994line} are two popular methods which use temporal difference algorithms. The policy-based approach is to optimize the policy $\pi_{\theta}(a|s)$ directly and update the parameters $\theta$ by gradient descent \cite{li2017deep}.
\par
The actor-critic algorithm is a reinforcement learning method combining policy gradient and temporal differential learning, which learns both a policy and a state value function. It estimates the parameters of two structures simultaneously.
\begin{enumerate}[]
	\item The actor is a policy function, which is to learn a policy $\pi_{\theta}(a|s)$ to obtain the highest possible return.
	\item The critic refers to the learned value function $V_{\phi}(s)$, which estimates the value function of the current policy, that is to evaluate the quality of the actor.
\end{enumerate}
In the actor-critic method, the critic solves a problem of prediction, while the actor pays attention to the control \cite{bhatnagar2009natural}.
There is more information of actor-critic method in \cite{sutton2018reinforcement,bhatnagar2009natural}
\par
The summary of the value-based method, the policy-based method, and the actor-critic method are as follows:

\begin{enumerate}[]
	\item The value-based method: It needs to calculate value function, and usually gets a definite policy.
	\item The policy-based method: It optimizes the policy $\pi$ without selecting an action according to value function.
	\item The actor-critic method: It combines the above two methods, and learns both the policy $\pi$ and the state value function.
	
\end{enumerate}
\par

Deep reinforcement learning (DRL) combines reinforcement learning and deep learning, which defines problems and optimizes goals in the framework of RL, and solves problems such as state representation and strategy representation using deep learning techniques.

\par
DRL has achieved great success in many challenging control tasks and uses DNNs to represent the control policy. For neural network training, a simple stochastic gradient algorithm or other first-order algorithms are usually chosen, but these algorithms are not efficient in exploring the weight space, which makes DRL methods often take several days to train \cite{wu2017scalable}. So, a distributed method was proposed to solve this problem, in which parallel actor-learners have a stabilizing effect during training \cite{mnih2016asynchronous}. It executes multiple agents to interact with the environment simultaneously, which reduces the training time. But this method ignores the sampling efficiency. A scalable and sample-efficient
natural gradient algorithm was proposed, which uses a Kronecker-factored approximation method to compute the natural policy gradient update, and employ the update to the actor and the critic (ACKTR) \cite{wu2017scalable}.

\subsection{Optimization in Meta Learning}

Meta learning \cite{schmidhuber1987evolutionary, Schaul2010} is a popular research direction in the field of machine learning. It solves the problem of learning to learn. In the past cognition, the research of machine learning is to obtain a large amount of data in a specific task firstly and then use the data to train the model. In machine learning, adequate training data is the guarantee of achieving good performance. However, human beings can well process new tasks with only a few training samples, which are much more efficient than traditional machine learning methods. The key point could be that the human brain has learned ``how to learn'' and can make full use of past knowledge and experience to guide the learning of new tasks. Therefore, how to make machines have the ability to learn efficiently like human beings has become a frontier issue in machine learning.

The goal of meta learning is to design a model that can training well in the new tasks using as few samples as possible without overfitting.
The process of adapting to the new tasks is essentially a learning process in the meta-testing, but only with limited samples from new tasks.
The application of meta learning methods in supervised learning can solve the few-shot learning problems \cite{finn2017model}.

As few-shot learning problems receive more and more attention, meta learning is also developing rapidly. In general, meta learning methods can be summarized into the following three types \cite{ModelvsOptimizationMetaLearning}: metric-based methods \cite{bromley1994signature, koch2015siamese, vinyals2016matching, snell2017prototypical}, model-based methods \cite{santoro2016meta, weston2014memory} and optimization-based methods \cite{andrychowicz2016learning, ravi2016optimization, finn2017model}. In this subsection, we focus on the optimization-based meta learning methods. In meta learning, there are usually some tasks with sufficient training samples and a new task with only a few training samples. The main idea can be described as follows: in the meta-train step, sample a task $\tau$ from the total task set $\mathcal{T}$, which contains $(D_\tau^{train}, D_\tau^{test})$. For task $\tau$, train and update the optimizer parameter $\theta$ with the training samples $D_\tau^{train}$, update the meta-optimizer parameter $\phi$ with the test samples $D_\tau^{test}$. The process of sampling tasks and updating parameters are repeated multiple times. In the meta-test step, the trained meta-optimizer is used for learning a new task.

Since the purpose of meta learning is to achieve fast learning, a key point is to make the gradient descent more accurately in the optimization. In some meta learning methods, the optimization process itself can be regarded as a learning problem to learn the prediction gradient rather than a determined gradient descent algorithm \cite{thrun2012learning}. Neural networks with original gradient as input and prediction gradient as output is often used as a meta-optimizer \cite{andrychowicz2016learning}. The neural work is trained using the training and test samples from other tasks and used in the new task. The parameter update in the process of training is as follows:

\begin{equation}
\theta_{t+1}=\theta_t+N(g(\theta_t),\phi),
\end{equation}

\noindent
where $\theta_t$ is the model parameter at the iteration $t$, and $N$ is the meta-optimizer with parameter $\phi$ that learns how to predict the gradient. After training, the meta-optimizer $N$ and its parameter $\phi$ are updated according to the loss value in the test samples. The experiments have confirmed that learning neural optimizers is advantageous compared to the most advanced adaptive stochastic gradient optimization methods used in deep learning \cite{andrychowicz2016learning}.
Due to the similarity between the gradient update in backpropagation and the cell state update in the long short-term memory (LSTM), LSTM is often used as the meta-optimizer \cite{ andrychowicz2016learning, ravi2016optimization}.

A model-agnostic meta learning algorithm (MAML) is another method for meta learning which was proposed to learn the parameters of any model subjected to gradient descent methods. It is applicable to different learning problems, including classification, regression and reinforcement learning \cite{ finn2017model}. The basic idea of the model-agnostic algorithm is to begin multiple tasks at the same time, and then get the synthetic gradient direction of different tasks, so as to learn a common base model.
The main process can be described as follows: in the meta-train step, multiple tasks batch $\tau_i$, which contains $(D^{train}_i, D^{test }_i)$, are extracted from the total task set $\mathcal{T}$. For all $\tau_i$, train and update the parameter $\theta_i^{'}$ with the train samples $D^{train}_i$:

	\begin{equation}
	\theta_i^{'}=\theta-\alpha \frac{\partial J_{\tau_i}(\theta)}{\partial (\theta)},
	\end{equation}
where $\alpha$ is the learning rate of training process and $J_{\tau_i}(\theta)$ is the loss function in task $i$ with training samples $D_i^{train}$. After the training step, use the synthetic gradient direction of these parameters $\theta_i^{'}$ on the test samples $D^{test}_i$ of the respective task to update parameter $\theta$:
\begin{equation}
\theta =\theta -\beta \frac{\partial \sum_{\tau_i \sim p(\mathcal{T})}J_{\tau_i}(\theta_i^{'})}{\partial (\theta)},
\end{equation}
where $\beta$ is the meta learning rate of the test process and $J_{\tau_i}(\theta)$ is the loss function in task $i$ with test samples $D_i^{test}$. The meta-train step is repeated multiple times to optimize a good initial parameter $\theta$. In the meta-test step, the trained parameter $\theta$ is used as the initial parameter such that the model has a maximal performance on the new task.
MAML does not introduce additional parameters for meta learning, nor does it require a specific learner architecture. The development of the method is of great significance to the optimization-based meta learning methods. Recently, an expanded task-agnostic meta learning algorithm is proposed to enhance the generalization of meta-learner towards a variety of tasks, which achieves outstanding performance on few-shot classification and reinforcement learning tasks \cite{Jamal_2019_CVPR}.

\subsection{Optimization in Variational Inference}
In the machine learning community, there are many attractive probabilistic models but with complex structures and intractable posteriors, and thus some approximate methods are used, such as variational inference and Markov chain Monte Carlo (MCMC) sampling. Variational inference, a common technique in machine learning, is widely used to approximate the posterior density of the Bayesian model, which transforms intricate inference problems into high-dimensional optimization problems \cite{jordan1999introduction,wainwright2008graphical}. Compared with MCMC, the variational inference is faster and more suitable for dealing with large-scale data. Variational inference has been applied to large-scale machine learning tasks, such as large-scale document analysis, computer vision and computational neuroscience \cite{blei2017variational}.

\par
Variational inference often defines a flexible family of distributions indexed by free parameters on latent variables \cite{jordan1999introduction}, and then finds the variational parameters by solving an optimization problem.

Now let us review the principle of variational inference \cite{hoffman2013stochastic}. Variational inference approximates the true posterior by attempting to minimize the Kullback-Leibler (KL) divergence between a potential factorized distribution and the true posterior.
\par
Let $Z=\{z_i\}$ represent the set of all latent variables and parameters in the model and $X=\{x_i\}$ be a set of all observed data. The joint likelihood of $X$ and $Z$ is $p(Z,X)=p(Z)p(X|Z) $. In Bayesian models, the posterior distribution $p(Z|X)$ should be computed to make further inference.
\par
What we need to do is to approximate $p(Z|X)$ with the distribution $q(Z)$ that belongs to a constrained family of distributions. The goal is to make the two distributions as similar as possible. Variational inference chooses KL divergence to measure the difference between the two distributions, that is to minimize the KL divergence of $q(Z)$ and $p(Z|X)$. Here is the formula for the KL divergence between $q$ and $p$:
\begin{align}
&\mathrm{KL}[q(Z)||p(Z|X)]=\mathbb{E}_q\left[\log\frac{q(Z)}{p(Z|X)}\right] \notag \\
&=\mathbb{E}_q[\log q(Z)]-\mathbb{E}_q[\log p(Z|X)] \notag \\
&=\mathbb{E}_q[\log q(Z)]-\mathbb{E}_q[\log p(Z,X)]+\log p(X) \notag \\
&=-\mathrm{ELBO}(q)+const,
\end{align}
where $\log p(X)$ is replaced by a constant because we are only interested in $q$.
With the above formula, we can know KL divergence is difficult to optimize because it requires knowing the distribution that we are trying to approximate. An alternative method is to maximize the evidence lower bound ($\mathrm{ELBO}$), a lower bound on the logarithm of the marginal probability of the observations. We can obtain $\mathrm{ELBO}$'s formula as
\begin{align}
\mathrm{ELBO}(q)=\mathbb{E}\left[\log p(Z,X)\right]-\mathbb{E}\left[\log q(Z)\right].
\end{align}
\par
Variational inference can be treated as an optimization problem with the goal of minimizing the evidence lower bound. A direct method is to solve this optimization problem using the coordinate ascent, which is called coordinate ascent variational inference (CAVI). CAVI iteratively optimizes each factor of the mean-field variational density, while holding the others fixed \cite{blei2017variational}.
\par
Specifically, variational distribution $q$ has the structure of the mean-field, i.e., $q(Z)=\prod_{i=1}^Mq_i(z_i)$. With this assumption, we can bring the distribution $q$ into the $\mathrm{ELBO}$, by some derivation according to \cite{Bishop2006Pattern}, and obtain the following formula:
\begin{equation}
q_i^*\propto \exp\{\mathbb{E}_{-i}[\log p(z_i,Z_{-i},X)]\}.
\end{equation}
Then the CAVI algorithm can be given below in Algorithm \ref{al:CAVI}.
\begin{algorithm}
	\caption{Coordinate Ascent Variational Inference \cite{blei2017variational}}\label{al:CAVI}
	\begin{algorithmic} 
		\Require ~~$p(X,Z)$, $X$
		\Ensure~~ $q(Z)=\prod_{i=1}^M q_i(z_i)$\\
		Initialize Variational factors $q_i(z_i)$
		\Repeat
		\For{i=1,2,3....,M}
		\State $q_i^*\propto \exp\{\mathbb{E}_{-i}[\log p(z_i,Z_{-i},X)]\}$
		\EndFor
		\State Compute $\mathrm{ELBO}(q)$:
		$$\mathrm{ELBO}(q)=\mathbb{E}[\log p(Z,X)]-\mathbb{E}\log q(Z)$$
		\Until{$\mathrm{ELBO}$ converges}
	\end{algorithmic}
\end{algorithm}

In traditional coordinate ascension algorithms, the efficiency of processing large data is very low, because each iteration needs to compute all the data, which is very time-consuming. Modern machine learning models often need to analyze and process large-scale data, which is difficult and costly. Stochastic optimization enables machine learning to be extended on massive data \cite{bottou2004large}. This reminds us of an attractive technique to handle large data sets: stochastic optimization \cite{robbins1951textordfemininea,blei2017variational,spall2005introduction}. By introducing stochastic optimization into variational inference, the stochastic variational inference (SVI) was proposed \cite{hoffman2013stochastic}, in which the exponential family is taken as a typical example.
\par
Gaussian process (GP) is an important machine learning method based on statistical learning and Bayesian theory. It is suitable for complex regression problems such as high dimensions, small samples, and nonlinearities. GP has the advantages of strong generalization ability, flexible non-parametric inference, and strong interpretability. However, the complexity and storage requirements of accurate solution for GP are high, which hinders the development of GP under large-scale data.
The stochastic variational inference method introduced in this section can popularize variational inference on large-scale datasets, but it can only be applied to probabilistic models with factorized structures. For GPs whose observations are correlated with each other, the stochastic variational inference can be adapted by introducing the global inducing variables as variational variables \cite{hensman2013gaussian, hensman2015scalable}. Specifically, the observations are assumed to be conditionally independent given the inducing variables and the variational distribution for the inducing variables is assumed to have an explicit form. Thus, the resulting GP model can be factorized in a necessary manner, enabling the stochastic variational inference. This method can also be easily extended to models with non-Gaussian likelihood or latent variable models based on GPs.

\subsection{Optimization in Markov Chain Monte Carlo}
Markov chain Monte Carlo (MCMC) is a class of sampling algorithms to simulate complex distributions that are difficult to sample directly. It is a practical tool for Bayesian posterior inference. The traditional and common MCMC algorithms include Gibbs sampling, slice sampling, Hamiltonian Monte Carlo (HMC) \cite{duane1987hybrid,neal2011mcmc}, Reimann manifold variants \cite{girolami2011riemann}, and so on. These sampling methods are limited by the computational cost and are difficult to extend to large-scale data.This section takes HMC as an example to introduce the optimization in MCMC. The bottleneck of the HMC is that the gradient calculation is costly on large data sets.
\par
We first introduce the derivation of HMC. Consider the random variable $\theta$, which can be sampled from the posterior distribution,
\begin{equation}
p(\theta|D)\propto \exp(-U(\theta)),
\end{equation}
where $D$ is the set of observations, and $U$ is the potential energy function with the following formula:
\begin{equation}
U(\theta)=-\log p(\theta|D)=-\sum_{x\in D}\log p(x|\theta)-\log p(\theta).
\end{equation}
In HMC \cite{duane1987hybrid}, an independent auxiliary momentum variable $r$ is introduced from Hamiltonian dynamic. The Hamiltonian function and the joint distribution of $\theta$ and $r$ are described by
\begin{equation}
H(\theta,r)=U(\theta)+\frac{1}{2}r^{T}M^{-1}r=U(\theta)+K(r),
\end{equation}
\begin{equation}
p(\theta,r)\propto \exp(-U(\theta)-\frac{1}{2}r^{T}M^{-1}r),
\end{equation}
where $M$ denotes the mass matrix, and $K(r)$ is the kinetic energy function. The process of HMC sampling is derived by simulating the Hamiltonian dynamic system,

\begin{equation}
\left\{
\begin{array}{lr}
d\theta = M^{-1}rdt, &  \\
dr = -\nabla U(\theta)dt. &
\end{array}
\right.
\end{equation}
Hamiltonian dynamic describes the continuous motion of a particle. Hamiltonian equations are numerically approximated by the discretized leapfrog integrator for practical simulating \cite{duane1987hybrid}. The update equations are as follows \cite{duane1987hybrid}:
\begin{equation}
\left\{
\begin{array}{lr}
r_i(t+\frac{\epsilon}{2})=r_i(t)-\frac{\epsilon}{2}dr(t), &  \\
\theta_i(t+\epsilon)=\theta_i(t)+\epsilon d\theta(t+\frac{\epsilon}{2}) ,&  \\
r_i(t+\epsilon)=r_i(t+\frac{\epsilon}{2})-\frac{\epsilon}{2}dr(t+\epsilon).
\end{array}
\right.
\end{equation}

\par
In the case of large datasets, the gradient of $U(\theta)$ needs to be calculated on the entire data set in each leapfrog iteration.
In order to improve the efficiency, the stochastic gradient method was used to calculate $\nabla U(\theta)$ with a mini-batch $\tilde{D}$ sampled uniformly from $D$, which reduces the cost of calculation \cite{chen2014stochastic}. However, the gradient calculated in a mini-batch instead of the full dataset will cause noise. According to the central limit theorem, this noisy gradient can be approximated as
\begin{equation}
\nabla \tilde{U}(\theta)\approx \nabla U(\theta)+ \mathcal N(0,V(\theta)),
\end{equation}

\noindent
where gradient noise obeys normal distribution whose covariance is $V(\theta)$.
If we replace $\nabla {U}(\theta)$ by $\nabla \tilde{U}(\theta)$ directly, the Hamiltonian dynamics will be changed as
\begin{equation}
\left\{
\begin{array}{lr}
d\theta = M^{-1}rdt ,&  \\
dr = -\nabla U(\theta)dt+\mathcal N(0,2B(\theta)dt) ,&
\end{array}
\right.
\end{equation}
where $B(\theta)=\frac{1}{2}\epsilon V(\theta)$ is the diffusion matrix \cite{chen2014stochastic}.

\par
Since the discretization of the dynamical system introduces noise, the Metropolis-Hastings (MH) correction step should be done after the leapfrog step. These MH steps require expensive calculations overall data in each iteration.
Beyond that, there is an incorrect stationary distribution \cite{betancourt2015fundamental} in the stochastic gradient variant of HMC.
Thus, Hamiltonian dynamic was further modified, which minimizes the effect of the additional noise, achieves the invariant distribution and eliminates MH steps \cite{chen2014stochastic}.
Specifically, a friction term is added to the dynamical process of momentum update:
\begin{equation}
\left\{
\begin{array}{lr}
d\theta = M^{-1}rdt, &  \\
dr = -\nabla U(\theta)dt-BM^{-1}rdt+\mathcal N(0,2B(\theta)dt).&
\end{array}
\right.
\end{equation}
The introduced friction term is helpful for decreasing total energy $H(\theta,r)$ and weakening the effects of noise in the momentum update phase.
The dynamical system is also the type of second-order Langevin dynamics with friction in physics, which can explore efficiently and counteract the effect of the noisy gradients \cite{chen2014stochastic} and thus no MH correction is required. This second-order Langevin dynamic MCMC method, called SGHMC, is used to deal with sampling problems on large data sets \cite{chen2014stochastic,ahn2012bayesian}.

Moreover, HMC is highly sensitive to hyper-parameters, such as the path length (step number) $L$ and the step size $\epsilon$. If the hyper-parameters are not set properly, the efficiency of the HMC will drop dramatically. There are some methods to optimize these two hyper-parameters instead of manually setting them.

\subsubsection{Path Length $L$}

The value of path length $L$ has a great influence on the performance of HMC. If $L$ is too small, the distance between the resulting sample points will be very close; if $L$ is too large, the resulting sample points will loop back, resulting in wasted computation. In general, manually setting $L$ cannot maximize the sampling efficiency of the HMC.

\par
Matthew et al. \cite{hoffman2014no} proposed an extension of the HMC method called the No-U-Turn sampler (NUTS), which uses a recursive algorithm to generate a set of possible independent samples efficiently, and stops the simulation by discriminating the backtracking automatically. There is no need to set the step parameter $L$ manually. In models with multiple discrete variables, the ability of NUTS to select the track length automatically allows it to generate more valid samples and perform more efficiently than the original HMC.

\subsubsection{Adaptive Step Size $\epsilon$}

The performance of HMC is highly sensitive to the step size $\epsilon$ in leapfrog integrator. If $\epsilon$ is too small, the update will slow, and the calculation cost will be high; if $\epsilon$ is too large, the rejection rate will be high, resulting in useless updates.
\par
To set $\epsilon$ reasonably and adaptively, a vanishing adaptation of the dual averaging algorithm can be used in HMC \cite{nesterov2009primal, andrieu2008tutorial}.
Specifically, a statistic $H_t=\delta-\alpha_t$ is adopted in dual averaging method, where $\delta$ is the desired average acceptance probability, and $\alpha_t$ is the current Metropolis-Hasting acceptance probability for iteration $t$. The statistic $H_t$'s expectation $h(\epsilon)$ is defined as

\begin{equation}
h(\epsilon)\equiv \mathbb{E}_t[H_t|\epsilon_t]\equiv \lim_{T\rightarrow \infty }\frac{1}{T}\mathbb{E}[H_t|\epsilon_t],
\end{equation}
where $\epsilon_t$ is the step size for iteration $t$ in the leapfrog integrator. To satisfy $h(\epsilon)\equiv \mathbb{E}_t[H_t|\epsilon_t]=0$, we can derive the update formula of $\epsilon$, i.e., $\epsilon_{t+1} = \epsilon_{t}-\eta_tH_t$.
Tuning $\epsilon$ by vanishing adaptation algorithm guarantees that the average acceptance probability of Metropolis verges to a fixed value.

\par
The hyper-parameters in the HMC include not only the step size $\epsilon$ and the length of iteration steps $L$, but also the mass $M$, etc. Optimizing these hyper-parameters can help improve sampling performance \cite{girolami2011riemann,carpenter2017stan,cotter2013mcmc}. It is convenient and efficient to tune the hyper-parameters automatically without cumbersome adjustments based on data and variables in MCMC. These adaptive tuning methods can be applied to other MCMC algorithms to improve the performance of the samplers.

\par
In addition to second-order SGHMC, stochastic gradient Langevin dynamics (SGLD) \cite{welling2011bayesian} is a first-order Langevin dynamic technique combined with stochastic optimization. Efficient variants of both SGLD and SGHMC are still active \cite{ahn2012bayesian,ding2014bayesian}.

\section{Challenges and Open Problems}\label{sec:challenges}
With the rise of practical demand and the increase of the complexity of machine learning models, the optimization methods in machine learning still face challenges. In this part, we discuss open problems and challenges for some optimization methods in machine learning, which may offer suggestions or ideas for future research and promote the wider application of optimization methods in machine learning.

\subsection{Challenges in Deep Neural Networks}
There are still many challenges while optimizing DNNs. Here we mainly discuss two challenges with respect to data and model, respectively. One is insufficient data in training, and the other is a non-convex objective in DNNs.

\subsubsection{Insufficient Data in Training Deep Neural Networks}

\par
In general, deep learning is based on big data sets and complex models. It requires a large number of training samples to achieve good training effects. But in some particular fields, finding a sufficient amount of training data is difficult. If we do not have enough data to estimate the parameters in the neural networks, it may lead to high variance and overfitting.
\par
There are some techniques in neural networks that can be used to reduce the variance. Adding $L_2$ regularization to the objective is a natural method to reduce the model complexity. Recently, a common method is dropout \cite{srivastava2014dropout}. In the training process, each neuron is allowed to stop working with a probability of $p$, which can prevent the synergy between certain neurons. $M$ subnets can be sampled like bagging by multiple puts and returns \cite{breiman1996bagging}. Each expected result at the output layer is calculated as
\begin{equation}
o=\mathbb{E}_M[f(x;\theta,M)]=\sum_{i=1}^M p(M_i)f(x;\theta,M_i),
\end{equation}
where $p(M_i)$ is the probability of the $i$th subnet. Dropout can prevent overfitting and improve the generalization ability of the network, but its disadvantage is increasing the training time as each training changes from the full network to a sub-network \cite{zaremba2014recurrent}.

\par

Not only overfitting but also some training details will affect the performance of the model due to the complexity of the DNNs. The improper selection of the learning rate and the number of iterations in the SGD will make the model unable to converge, which makes the accuracy of model fluctuate greatly. Besides, taking an inappropriate black box of neural network construction may result in training not being able to continue, so designing an appropriate neural network model is particularly important. These impacts are even greater when data are insufficient.

\par
The technology of transfer learning \cite{pan2010survey} can be applied to build networks in the scenario of insufficient data. Its idea is that the models trained from other data sources can be reused in similar target fields after certain modifications and improvements, which dramatically alleviates the problems caused by insufficient datasets. Moreover, the advantages brought by transfer learning are not limited to reducing the need for sufficient training data, but also can avoid overfitting effectively and achieve better performance in general.
However, if target data is not as relevant to the original training data, the transferred model does not bring good performance.

Meta learning methods can be used for systematically learning parameter initialization, which ensures that training begins with a suitable initial model. However, it is necessary to ensure the correlation between multiple tasks for meta-training and tasks for meta-testing.
Under the premise of models with similar data sources for training, transfer learning and meta learning can overcome the difficulties caused by insufficient training data in new data sources, but these methods usually introduce a large number of parameters or complex parameter adjustment mechanisms, which need to be further improved for specific problems. Therefore, using insufficient data for training DNNs is still a challenge.

\subsubsection{Non-convex Optimization in Deep Neural Network}
Convex optimization has good properties and a comprehensive set of tools are open to solve the optimization problem. However, many machine learning problems are formulated as non-convex optimization problems. For example, almost all the optimization problems in DNNs are non-convex. Non-convex optimization is one of the difficulties in the optimization problem. Unlike convex optimization, there may be innumerable optimum solutions in its feasible domain in non-convex problems. The complexity of the algorithm for searching the global optimal value is NP-hard \cite{allen2016variance}.

\par
In recent years, non-convex optimization has gradually attracted the attention of researches. The methods for solving non-convex optimization problems can be roughly divided into two types. One is to transform the non-convex optimization into a convex optimization problem, and then use the convex optimization method. The other is to use some special optimization method for solving non-convex functions directly. There is some work on summarizing the optimization methods for solving non-convex functions from the perspective of machine learning \cite{jain2017non}.
\begin{enumerate}
    \item Relaxation method: Relax the problem to make it become a convex optimization problem. There are many relaxation techniques, for example, the branch-and-bound method called $\alpha$BB convex relaxation \cite{adjiman1998global,adjiman1998mixed}, which uses a convex relaxation at each step to compute the lower bound in the region.
	The convex relaxation method has been used in many fields. In the field of computer vision, a convex relaxation method was proposed to calculate minimal partitions \cite{pock2009convex}. For unsupervised and semi-supervised learning, the convex relaxation method was used for solving semidefinite programming \cite{xu2005unsupervised}.
\item Non-convex optimization methods: These methods include projection gradient descent \cite{chen2015fast,PNCPGD}, alternating minimization \cite{jain2013low,hardt2014understanding,hardt2014fast}, expectation maximization algorithm \cite{balakrishnan2017statistical,wang2014high} and stochastic optimization and its variants \cite{johnson2013accelerating}.
\end{enumerate}
\par

\subsection{Difficulties in Sequential Models with Large-Scale Data}
When dealing with large-scale time series, the usual solutions are using stochastic optimization, processing data in mini-batches, or utilizing distributed computing to improve computational efficiency \cite{keskar2016large}.
For a sequential model, segmenting the sequences can affect the dependencies between the data on the adjacent time indices.
If sequence length is not an integral multiple of the mini-batch size, the general operation is to add some items sampled from the previous data into the last subsequence. This operation will introduce the wrong dependency in the training model. Therefore, the analysis of the difference between the approximated solution obtained and the exact solution is a direction worth exploring.
\par
Particularly, in RNNs, the problem of gradient vanishing and gradient explosion is also prone to occur. So far, it is generally solved by specific interaction modes of LSTM and GRU \cite{chung2014empirical} or gradient clipping. Better appropriate solutions for dealing with problems in RNNs are still worth investigating.

\subsection{High-Order Methods for Stochastic Variational Inference}
The high-order optimization method utilizes curvature information and thus converges fast. Although computing and storing the Hessian matrices are difficult, with the development of research, the calculation of the Hessian matrix has made great progress \cite{roosta2016sub,xu2016sub,martens2016second}, and the second-order optimization method has become more and more attractive.
Recently, stochastic methods have also been introduced into the second-order method, which extends the second order method to large-scale data \cite{roosta2016sub,bollapragada2016exact}.

\par
We have introduced some work on stochastic variational inference. It introduces the stochastic method into variational inference, which is an interesting and meaningful combination. This makes variational inference be able to handle large-scale data. A natural idea is whether we can incorporate second-order optimization methods (or higher-order) into stochastic variational inference, which is interesting and challenging.

\subsection{Stochastic Optimization in Conjugate Gradient}
Stochastic methods exhibit powerful capabilities when dealing with large-scale data, especially for first-order optimization \cite{schraudolph2002conjugate}. Then the relevant experts and scholars also introduced this stochastic idea to the second-order optimization methods \cite{schraudolph2007stochastic,byrd2016stochastic,bordes2009sgd} and achieved good results.
\par
Conjugate gradient method is an elegant and attractive algorithm, which has the advantages of both the first-order and second-order optimization methods. The standard form of a conjugate gradient is not suitable for a stochastic approximation. Through using the fast Hessian-gradient product, the stochastic method is also introduced to conjugate gradient, in which some numerical results show the validity of the algorithm \cite{schraudolph2002conjugate}. Another version of stochastic conjugate gradient method employs the variance reduction technique, and converges quickly with just a few iterations and requires less storage space during the running process \cite{jin2018stochastic}. The stochastic version of conjugate gradient is a potential optimization method and is still worth studying.

\section{Conclusion}\label{sec:conclusion}
This paper introduces and summarizes the frequently used optimization methods from the perspective of machine learning, and studies their applications in various fields of machine learning. Firstly, we describe the theoretical basis of optimization methods from the first-order, high-order, and derivative-free aspects, as well as the research progress in recent years. Then we describe the applications of the optimization methods in different machine learning scenarios and the approaches to improve their performance. Finally, we discuss some challenges and open problems in machine learning optimization methods.

\bibliographystyle{IEEEtran}

\bibliography{mybibfile}

%





\end{document}